\documentclass[11pt,a4paper]{article}
\usepackage[hyperref]{acl2021}
\usepackage{times}
\usepackage{latexsym}

\usepackage{CJK}
\usepackage{color}
\usepackage{array}
\usepackage{amsmath,bm}
\usepackage{amsthm}
\usepackage{graphicx} 
\usepackage{mathrsfs}
\usepackage{amsfonts,amssymb}
\usepackage{algorithm}
\usepackage[noend]{algpseudocode}
\usepackage{setspace}
\usepackage{multirow}
\usepackage{graphicx}
\usepackage{subcaption}
\usepackage[shortlabels]{enumitem}
\setlist[enumerate]{nosep}
\def\infinity{\rotatebox{90}{8}}
\algrenewcommand\algorithmicindent{1.2em}%
\usepackage{microtype}
\usepackage{amsmath}
\usepackage{mathtools}
\usepackage{amssymb}
\usepackage{hyperref}

\let\emptyset\varnothing
\DeclareMathOperator*{\argmax}{arg\,max}

\aclfinalcopy 
\def\aclpaperid{1104} 

\newcommand{\norm}[1]{\left\lVert #1 \right\rVert}

\newtheorem{theorem}{Theorem}
\newtheorem{corollary}{Corollary}[theorem] 

\title{Certified Robustness to Text Adversarial Attacks by Randomized [MASK]}

\author{Jiehang Zeng, Xiaoqing Zheng\thanks{\enspace Corresponding author} \,, Jianhan Xu, \\ \textbf{Linyang Li, Liping Yuan, Xuanjing Huang} \\
School of Computer Science, Fudan University, Shanghai, China\\
 \texttt{\{jhzeng18, zhengxq, jianhanxu20\}@fudan.edu.cn } \\
\texttt{\{linyangli19, lpyuan19, xjhuang\}@fudan.edu.cn} \\}

\date{}

\begin{document}
\begin{CJK}{UTF8}{gbsn}

\maketitle
\begin{abstract}
Recently, few certified defense methods have been developed to provably guarantee the robustness of a text classifier to adversarial synonym substitutions. 
However, all existing certified defense methods assume that the defenders are informed of how the adversaries generate synonyms, which is not a realistic scenario.
In this paper, we propose a certifiably robust defense method by randomly masking a certain proportion of the words in an input text, in which the above unrealistic assumption is no longer necessary.
The proposed method can defend against not only word substitution-based attacks, but also character-level perturbations.
We can certify the classifications of over $50\%$ texts to be robust to any perturbation of five words on AGNEWS, and two words on SST2 dataset.
The experimental results show that our randomized smoothing method significantly outperforms recently proposed defense methods across multiple datasets.



\end{abstract}

\section{Introduction}
 
Although deep neural networks have achieved prominent performance on many natural language processing (NLP) tasks, they are vulnerable to adversarial examples that are intentionally crafted by replacing, scrambling, and erasing characters \cite{gao2018black,ebrahimi-etal-2018-hotflip} or words \cite{alzantot2018generating,ren-etal-2019-generating, zheng2020evaluating, Jin_Jin_Zhou_Szolovits_2020, li2020bert} under certain semantic and syntactic constraints.
These adversarial examples are imperceptible to humans but can easily fool deep neural network-based models.
The existence of adversarial examples has raised serious concerns, especially when deploying such NLP models to security-sensitive tasks. 
In Table \ref{table:textual_adversarial_examles}, we list three examples that are labeled as positive sentiment by a BERT-based sentiment analysis model \cite{devlin2018bert}, and for each example, we give two adversarial examples (one is generated by character-level perturbations, and another is crafted by adversarial word substitutions) that can cause the same model to change its prediction from correct (positive) sentiment to incorrect (negative) one.

\begin{table*}[htbp]
\centering
\small
\begin{tabular}{l|l}
\hline
\hline
\multicolumn{1}{c|}{\textbf{Original example (positive)}} & \multicolumn{1}{c}{\textbf{Adversarial example (negative)}} \\
\hline
The tenderness of the piece is still intact. & The \textcolor{red}{\textbf{tendePness}} of the piece is still \textcolor{red}{\textbf{inTact}}. \\ 
 & The tenderness of the piece is still \textcolor{red}{\textbf{untouched}} . \\ 
\hline
A worthwhile way to spend two hours. & A \textcolor{red}{\textbf{w0rthwhile}} way to spend two hours. \\
& A worthwhile way to \textcolor{red}{\textbf{expend}} two hours. \\ 
\hline
Season grades for every college football team. & Season grades for every college \textcolor{red}{\textbf{fo0tba1l}} team. \\
& Season \textcolor{red}{\textbf{scores}} for every college football team. \\
\hline
\hline
\end{tabular}
\caption{Three example sentences and their adversarial examples. For each example, we list two adversarial examples. The first one is generated by character-level perturbations and the second is crafted by adversarial word substitutions. 
Those adversarial examples can successfully fool a BERT-based sentiment analysis model \cite{devlin2018bert} to classify them as negative sentiment while their original examples are label as positive sentiment by the same model. The words perturbed are highlighted in bold red ones.}
\label{table:textual_adversarial_examles}
\end{table*}

Many methods have been proposed to defend against adversarial attacks for neural natural language processing (NLP) models. Most of them are evaluated empirically, such as adversarial data augmentation \cite{Jin_Jin_Zhou_Szolovits_2020,zheng2020evaluating}, adversarial training \cite{madry2018towards, zhu2019freelb}, and Dirichlet Neighborhood Ensemble (DNE) \cite{zhou2020defense}.
Among them, adversarial data augmentation is one of the widely used methods \cite{Jin_Jin_Zhou_Szolovits_2020, zheng2020evaluating,li2020bert}.
During the training time, they replace a word with one of its synonyms that maximizes the prediction loss.
By augmenting these adversarial examples with the original training data, the model is robust to such perturbations.
However, it is infeasible to explore all possible combinations in which each word in a text can be replaced with any of its synonyms.

\citet{zhou2020defense} and \citet{iclr-21:Dong} relax a set of discrete points (a word and its synonyms) to a convex hull spanned by the word embeddings of all these points, and use a convex hull formed by a word and its synonyms to capture word substitutions. 
During the training, they samples an embedding vector in the convex hull to ensure the robustness within such a region.
To deal with complex error surface, a gradient-guided optimizer is also applied to search for more valuable adversarial points within the convex hull. 
By training on these virtual sentences, the model can enhance the robustness against word substitution-based perturbations.

Although the above-mentioned methods can empirically defend against the attack algorithms used during the training, the trained model often cannot survive from other stronger attacks \cite{Jin_Jin_Zhou_Szolovits_2020, li2020bert}.
A certified robust NLP model is necessary in both theory and practice. 
A model is said to be certified robust when it is guaranteed to give the correct answer under any attacker, no matter the strength of the attacker and no matter how the attacker manipulates the input texts.
Certified defense methods have recently been proposed~\cite{jia-etal-2019-certified,huang-etal-2019-achieving} by certifying the performance within the convex hull formed by the embeddings of a word and its synonyms.
However, due to the difficulty of propagating convex hull through deep neural networks, they compute a loose outer bound using Interval Bound Propagation (IBP).
As a result, IBP-based certified defense methods are hard to scale to large architectures such as BERT \cite{devlin2018bert}.

To achieve certified robustness on large architectures, 
\citet{ye-etal-2020-safer} proposed SAFER,
a randomized smoothing method that can provably ensure that the prediction cannot be altered by any possible synonymous word substitution.
However, existing certified defense methods assume that the defenders know how the adversaries generate synonyms, which is not a realistic scenario since we cannot impose a limitation on the synonym table used by the attackers.
In a real situation, we know nothing about the attackers and existing adversarial attack algorithms for NLP models may use a synonym table in which a single word can have many (up to $50$) synonyms \cite{Jin_Jin_Zhou_Szolovits_2020}, generate synonyms dynamically by using BERT \cite{li2020bert}, or perform character-level perturbations 
\cite{gao2018black,li2019textbugger} to launch adversarial attacks.

\begin{figure*}
 \centering
 \includegraphics[width = 13cm]{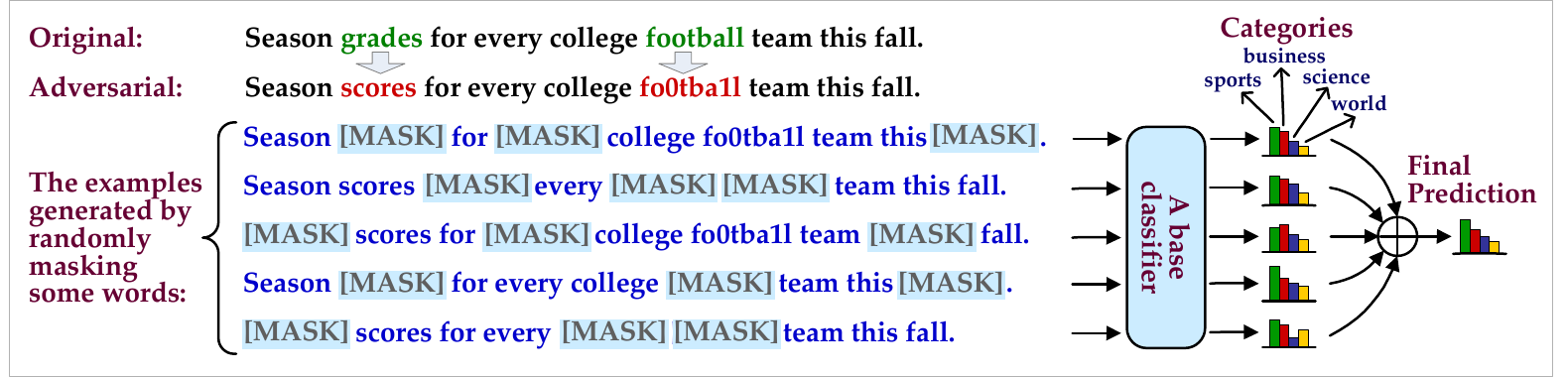}
 \caption{\label{fig:mask} Consider an original sentence given at the top, we assume that an adversarial example is created by replacing the word ``grades'' with ``scores'' and ``football'' with ``fo0tba1l''. Taking the adversarial example as input, we randomly mask three words (indicated by {\fontfamily{qcr}\selectfont [MASK]}) on the input and generate a set of masked copies. A base classifier is then used to label each of these masked copies (only five are shown here for clarity), and the prediction scores on the masked texts are ensembled to get a robust output. Our mask operation is the same as that used in training BERT \cite{devlin2018bert} or RoBERTa \cite{liu2019roberta} via masked language models, and the masked words can simply be encoded as the embedding of {\fontfamily{qcr}\selectfont [MASK]} so that we can leverage the ability of BERT to ``recover'' or ``reconstruct'' information about the masked words.}
\end{figure*}

In this paper, we propose \textbf{RanMASK}, a certifiably robust defense method against text adversarial attacks based on a new randomized smoothing technique for NLP models.
The proposed method works by repeatedly performing random ``mask'' operations on an input text, in order to generate a large set of masked copies of the text.  
A base classifier is then used to classify each of these masked texts, and the final robust classification is made by ``majority vote'' (see Figure \ref{fig:mask}).
In the training time, the base classifier is also trained with similar masked text samples.
Our mask operation is the same as that used in training BERT \cite{devlin2018bert} or RoBERTa \cite{liu2019roberta} via masked language models, and the masked words can simply be encoded as the embedding of {\fontfamily{qcr}\selectfont [MASK]} so that we can leverage the ability of BERT to ``recover'' or ``reconstruct'' information about the masked words.

The key idea is that, if a sufficient number of words are randomly masked from a text before the text is given to the base classifier and a relatively small number of words have been intentionally perturbed, then it is highly unlikely that all of the perturbed words (adversarially chosen) are present in a given masked text.
Note that remaining just some of these perturbed words is often not enough to fool the base classifier.
The results of our preliminary experiments also confirmed that textual adversarial attacks themselves are vulnerable to small random perturbations, and if we randomly mask few words from adversarial examples before they are fed into the classifier, it is more likely that the incorrect predictions of adversarial examples would be reverted to the correct ones.
Given a text $\boldsymbol{x}$ and a potentially adversarial text $\boldsymbol{x}'$, if we use a statistically sufficient number of random masked samples, and if the observed ``gap'' between the number of ``votes'' for the top class and the number of ``votes'' for any other class at $\boldsymbol{x}$ is sufficiently large, then we can guarantee with high probability that the robust classification at $\boldsymbol{x}'$ will be the same as it is at $\boldsymbol{x}$.
Therefore, we can prove that with high probability, the smoothed classifier will label $\boldsymbol{x}$ robustly against any text adversarial attack which is allowed to perturb a certain number of words in an input text at both of the word and character levels\footnote{This study was inspired by Levine and Feizi's work \cite{levine2020robustness} from the image domain, but our study is different from theirs in the key idea behind the method. In their proposed $\ell_0$ smoothing method, ``for each sample generated from $\boldsymbol{x}$, a \textbf{majority} of pixels are randomly dropped from the image before the image is given to the base classifier. If a relatively small number of pixels have been adversarially corrupted, then it is highly likely that \textbf{none} of these pixels are present in a given ablated sample. Then, for the majority of possible random ablations, $\boldsymbol{x}$ and $\boldsymbol{x}'$ will give the same ablated image.'' In contrast to theirs, our key idea is that, if a \textbf{sufficient} number of words are randomly masked from a text and a relatively small number of words have been intentionally perturbed, then it is highly unlikely that \textbf{all} of the perturbed words (adversarially chosen) are present in a given masked text. Remaining just some of these perturbed words is often not enough to fool the text classifier.}. 

The major advantage of our method over existing certified defense methods is that our certified robustness is not based on the assumption that the defenders know how the adversaries generate synonyms.
Given a text, the adversaries are allowed to replace few words with their synonyms (word-level perturbation) or deliberately misspell few of them (character-level perturbation).
Through extensive experiments on multiple datasets, we show that RanMASK achieved better performance on adversarial samples than existing defense methods. 
Experimentally, we can certify the classifications of over $50\%$ sentences to be robust to any perturbation of $5$ words on AGNEWS dataset \cite{zhang2015character}, and $2$ words on SST2 \cite{socher2013recursive}.
Furthermore, unlike most certified defenses (except SAFER), the proposed method is easy to implement and can be integrated into any existing neural network including those with large architecture such as BERT \cite{devlin2018bert}. 

Our contribution can be summarized as follows\footnote{The source code of RanMAKS is available on Github (\url{https://github.com/zjiehang/RanMASK})}:
\begin{itemize}
\item We propose RanMASK, a novel certifiably robust defense method against both the word substitution-based attacks and character-level perturbations.
The main advantage of our method is that we do not base the certified robustness on the unrealistic assumption that the defenders know how the adversaries generate synonyms.

\item We provide a general theorem that can be used to develop a related robustness certificate.
We can certify that the classifications of over $50\%$ sentences are robust to any modification of at most $5$ words on Ag's News Topic Classification Dataset (AGNEWS) \cite{zhang2015character}, and $2$ words on Stanford Sentiment Treebank Dataset (SST2) \cite{socher2013recursive}.

\item To further improve the empirical robustness, we proposed a new sampling strategy in which the probability of a word being masked corresponds to its output probability of a BERT-based language model (LM) to reduce the risk probability (see Section \ref{sec:empirical_result} for details). Through extensive experimentation we also show that our smoothed classifiers outperform existing empirical and certified defenses across different datasets.
\end{itemize}

\section{Preliminaries and Notation}

For text classification, a neural network-based classifier $f(\boldsymbol{x})$ maps an input text $\boldsymbol{x} \in \mathcal{X}$ to a label $y \in \mathcal{Y}$, where $\boldsymbol{x} = x_1, \dots, x_h$ is a text consisting of $h$ words and $\mathcal{Y}$ is a set of discrete categories. We follow mathematical notation used by Levine and Feizi \cite{levine2020robustness} below.

Given an original input $\boldsymbol{x}$, adversarial examples are created to cause a model to make mistakes.
The adversaries may create an adversarial example $\boldsymbol{x}' = x_1', \dots, x_h'$ by perturbing at most $d \leq h$ words in $\boldsymbol{x}$.
We say $\boldsymbol{x}'$ is a good adversarial example of $\boldsymbol{x}$ for untargeted attack if:
\begin{equation}
  f(\boldsymbol{x}') \neq y, \quad 
  \norm{\boldsymbol{x} - \boldsymbol{x}'}_0 \leq d
\end{equation}
\noindent where $y$ is the truth output for $\boldsymbol{x}$, and $\norm{\boldsymbol{x} - \boldsymbol{x}'}_0 = \sum_{i=1}^n{\mathbb{I}\{x_i \neq x_i'\}}$ is the Hamming distance, where $\mathbb{I}\{ \cdot \}$ is the indicator function. 
For character-level perturbations, $x'_i$ is a visually similar misspelling or typo of $x_i$, and for word-level substitutions, $x_i'$ is any of $x_i$'s synonyms, where the synonym sets are chosen by the adversaries, which usually may not be known by the defenders. 
A model $f$ is said to be \emph{certified robust} against adversarial attacks on an input $\boldsymbol{x}$ if it is able to give consistently correct predictions for all the possible adversarial examples $\boldsymbol{x}'$ under the constraint of $\norm{\boldsymbol{x} - \boldsymbol{x}'}_0 \leq d$. 

Let $\boldsymbol{x} \ominus \boldsymbol{x}'$ denote the set of word indices at which $\boldsymbol{x}$ and $\boldsymbol{x}'$ differ, so that $|\boldsymbol{x} \ominus \boldsymbol{x}'| = \norm{\boldsymbol{x} - \boldsymbol{x}'}_0$. 
For example, if $\boldsymbol{x} = $ ``{\fontfamily{qcr}\selectfont A B C E D}'' and $\boldsymbol{x}' = $ ``{\fontfamily{qcr}\selectfont A F C G D}'', $\boldsymbol{x} \ominus \boldsymbol{x}' = \{2, 4\}$ and $|\boldsymbol{x} \ominus \boldsymbol{x}'| = 2$.
Also, let $\mathcal{S}$ denote the set of indices $\{1, \dots, h\}$, $\mathcal{I}(h, k) \subseteq \mathcal{P}(\mathcal{S}) $ all possible sets of $k$ unique indices in $\mathcal{S}$, where $\mathcal{P}(\mathcal{S})$ is the power set of $\mathcal{S}$, and 
$\mathcal{U}(h, k)$ the uniform distribution over $\mathcal{I}(h, k)$. 
Note that to sample from $\mathcal{U}(h, k)$ is to sample $k$ out of $h$ indices uniformly without replacement.
For instance, an element sampled from $\mathcal{U}(5, 3)$ might be $\{1, 3, 5\}$.

We define a \emph{mask} operation $\mathcal{M}: \mathcal{X} \times \mathcal{I}(h, k) \rightarrow \mathcal{X}_{\text{mask}}$, where $\mathcal{X}_{\text{mask}}$ is a set of texts in which some words may be masked.
This operation takes an text of length $h$ and a set of indices as inputs and outputs the masked text, with all words except those in the set replaced with a special token {\fontfamily{qcr}\selectfont [MASK]}.
For example, $\mathcal{M}(\text{``{\fontfamily{qcr}\selectfont A F C G D}''}, \{1, 3, 5\}) = $ ``{\fontfamily{qcr}\selectfont A [MASK] C [MASK] D}''.
Following \citet{devlin2018bert}, we use {\fontfamily{qcr}\selectfont [MASK]} to replace the words masked.



\section{RanMASK: Certified Defense Method}
Inspired by the work of \cite{pmlr-v97-cohen19c, levine2020robustness} from the image domain, our method is to replace a base model $f$ with a more smoothed model that is easier to verify by ensembling the outputs of a number of random masked inputs.
In particular, let $f: \mathcal{X}_{\text{mask}} \rightarrow \mathcal{Y}$ be a base classifier, which is trained to classify texts with some words masked randomly, and a smoothed classifier $g(\boldsymbol{x})$ then can be defined as follows:
\begin{equation}
  \label{eq:smoothed_classifier}
  g(\boldsymbol{x}) = \argmax_{c \in \mathcal{Y}} \left[ \underset{\mathcal{H} \sim \mathcal{U}(h_{\boldsymbol{x}}, k_{\boldsymbol{x}})}{\mathbb{P}} (f(\mathcal{M}(\boldsymbol{x}, \mathcal{H})) = c) \right]
\end{equation}
\noindent where $h_{\boldsymbol{x}}$ is the length of $\boldsymbol{x}$, and $k_{\boldsymbol{x}}$ is the number of words retained (not masked) from $\boldsymbol{x}$, which is calculated by $\lfloor h_{\boldsymbol{x}} - \rho \times h_{\boldsymbol{x}} \rceil$, where $\rho$ is the percentage of words that can be masked.
To simplify notation, we let $p_c(\boldsymbol{x})$ denote the probability that, after randomly masking, $f$ returns the class $c$:
\begin{equation}
  \label{equ-pcx}
  p_c(\boldsymbol{x}) = \underset{\mathcal{H} \sim \mathcal{U}(h_{\boldsymbol{x}}, k_{\boldsymbol{x}})}{\mathbb{P}} (f(\mathcal{M}(\boldsymbol{x}, \mathcal{H})) = c).
\end{equation}
Therefore, $g(\boldsymbol{x})$ can be defined as $\argmax_{c \in \mathcal{Y}} p_c(\boldsymbol{x})$.
In other words, $g(\boldsymbol{x})$ denotes the class most likely to be returned if we first randomly mask all but $k_{\boldsymbol{x}}$ words from $\boldsymbol{x}$ and then classify the resulting text with the base classifier.

\subsection{Certified Robustness}

\begin{theorem}
\label{theorem-pcx-pcxpai-diff}
Given text $\boldsymbol{x}$ and $\boldsymbol{x}'$, $\norm{ \boldsymbol{x} - \boldsymbol{x}'}_0 \leq d$, for all class $c \in \mathcal{Y}$, we have:
\begin{equation}
 p_c(\boldsymbol{x}) - p_c(\boldsymbol{x}') \leq \beta \Delta
\end{equation}
where 
\begin{equation}
\label{equ-beta-definition}
\begin{gathered}
 \Delta = 1 - \frac{\tbinom{h_{\boldsymbol{x}} - d}{k_{\boldsymbol{x}}}}{\tbinom{h_{\boldsymbol{x}}}{k_{\boldsymbol{x}}}}, 
 \\
 \beta = \mathbb{P}(f(\mathcal{M}(\boldsymbol{x}, \mathcal{H})) = c \mid \mathcal{H} \cap (\boldsymbol{x} \ominus \boldsymbol{x}') \neq \emptyset).
 \end{gathered}
\end{equation}
\end{theorem}

The complete proof of Theorems \ref{theorem-pcx-pcxpai-diff} is given in Appendix \ref{sec:proof_of_theorem1}.
It is impossible to exactly compute the probabilities with which $f$ classifies $\mathcal{M}(\boldsymbol{x}, \mathcal{H})$ as each class.
Therefore, it is also impossible to exactly evaluate $p_c(\boldsymbol{x})$ and $g(\boldsymbol{x})$ at any input $\boldsymbol{x}$.
Let $\underline{p_c(\boldsymbol{x})}$ denote a lower bound on $p_c(\boldsymbol{x})$, we can bound 
$\underline{p_c(\boldsymbol{x})}$ with $(1 - \alpha)$ confidence.
Following \citet{pmlr-v97-cohen19c} and \citet{jia-etal-2019-certified}, we estimates $\underline{p_c(\boldsymbol{x})}$ using the standard one-sided Clopper-Pearson method \cite{clopper1934use}. Specifically, we randomly construct $n$ masked copies of $\boldsymbol{x}$, then count for class $c$ as $n_c = \sum_{1}^{n} {\mathbb{I}(f(\mathcal{M}(\boldsymbol{x}, \mathcal{H})) = c)}$ according to the outputs of $f$ (see Algorithm \ref{alg_overall_conclusion} for details). 
Assuming that $n_c$ follows a binomial distribution with parameters $n$ and $p_c$, $n_c \sim \text{B}(n, p_c)$, where $p_c = n_c / n$, we have:
\begin{equation}
\label{equ-lower-bound}
\underline{p_c(\boldsymbol{x})} = \text{Beta}(\alpha; n_c, n - n_c + 1),
\end{equation}
\noindent where $\text{Beta}(\alpha; u, v)$ is the $\alpha$-th quantile of a beta distribution with parameters $u$ and $v$, and $(1 - \alpha)$ is the confidence level.

\begin{corollary}
\label{corollary-certified}
For text $\boldsymbol{x}$, $\boldsymbol{x}'$, $\norm{ \boldsymbol{x} - \boldsymbol{x}'}_0 \leq d$, if 
\begin{equation}
\label{equ-certified-condition}
  \underline{p_c(\boldsymbol{x})} - \beta \Delta > 0.5
\end{equation}
then, with probability at least $(1 - \alpha)$:
\begin{equation}
  g(\boldsymbol{x}') = c
\end{equation}
\end{corollary}
\begin{proof}
With probability at least $(1 - \alpha)$ we have:
\begin{equation}
  0.5 < \underline{p_c(\boldsymbol{x})} - \beta \Delta \leq p_c(\boldsymbol{x}) - \beta \Delta \leq p_c(\boldsymbol{x}')
\end{equation}
\noindent Where the last inequality is from Theorem \ref{theorem-pcx-pcxpai-diff}, and $g(\boldsymbol{x}') = c$ from its definition given in Equation \eqref{eq:smoothed_classifier}. 
If $c = y$ (the truth output for $\boldsymbol{x}$) and if $\underline{p_c(\boldsymbol{x})} - \beta \Delta > 0.5$, 
the smoothed classifier $g$ is \emph{certified robust} at the input $\boldsymbol{x}$.
\end{proof}

Since the value of $\beta$ is always positive and less than $1$ by its definition, the condition \eqref{equ-certified-condition} is easier to be satisfied than the situation where the value of $beta$ is set to $1$ as \cite{levine2020robustness}, which yields a tighter certificate bound.
Note that Levine and Feizi \cite{levine2020robustness} assumes that all inputs are of equal length (i.e., the number of pixels is the same for all images) while we have to deal with variable-length texts here.
In this general theorem from which a related robustness certificate can be developed, we define the base classifier $f(\boldsymbol{x})$, the smoothed classifier $g(\boldsymbol{x})$, the values of $\Delta$ and $\beta$ based on a masking rate $\rho$ (i.e., the percentage of words that can be masked), while their counterparts are defined based on a retention constant, i.e, the fixed number of pixels retained from any input image.

\begin{algorithm*}[h]
\caption{For estimating the value of $\beta$}
\label{alg_estimating_beta}
\begin{algorithmic}[1]
\Procedure{BetaEstimator}{$\boldsymbol{x},h_{\boldsymbol{x}},k_{\boldsymbol{x}},r,f,n_r,n_k$}
  \State $\beta \gets 0$
  \State $\mathcal{A}$ $\gets$ Sampling $n_r$ elements from $\mathcal{U}(h_{\boldsymbol{x}},r)$
  \For{ each $a$ in $\mathcal{A}$ }
    \State $\mathcal{B}$ $\gets$ Sampling $n_k$ elements from $\mathcal{U}(h_{\boldsymbol{x}},k_{\boldsymbol{x}})$
    \For { each $b$ in $\mathcal{B}$ }
    \If {$a \cap b = \emptyset$} 
      \State $\mathcal{B}.delete(b)$
    \EndIf
    \EndFor
  \State $p_c \gets$ Using Eq. (\ref{equ-pcx}) with $f$ and $\mathcal{B}$.
  \State $\beta \gets \beta + p_c$
  \EndFor
\State $\beta \gets \beta / n_r$
\State \Return $\beta$
\EndProcedure
\end{algorithmic}
\end{algorithm*}

\subsection{Estimating the Value of Beta $\beta$}
We discuss how to estimate the value of $\beta$ defined in Theorem \ref{theorem-pcx-pcxpai-diff} here.
Recall that $\beta$ is the probability that $f$ will label the masked copies of $\boldsymbol{x}$ with the class $c$ where the indices of unmasked words are overlapped with $\boldsymbol{x} \ominus \boldsymbol{x}'$ (i.e., the set of word indices at which $\boldsymbol{x}$ and $\boldsymbol{x}'$ differ). 
We use a Monte Carlo algorithm to evaluating $\beta$ by sampling a large number of elements from $\mathcal{U}(h_{\boldsymbol{x}}, k_{\boldsymbol{x}})$.
To simplify notation, we let $r$ denote the value of $|\boldsymbol{x} \ominus \boldsymbol{x}'|$.

The Monte Carlo-based algorithm used to evaluate $\beta$ is given in Algorithm \ref{alg_estimating_beta}.
We first sample $n_r$ elements from $\mathcal{U}(h_{\boldsymbol{x}}, r)$ and each sampled element, denoted by $a$, is a set of indices where the words are supposed to be perturbed.
For every $a$, we then sample $n_k$ elements from $\mathcal{U}(h_{\boldsymbol{x}}, k_{\boldsymbol{x}})$, each of which, denoted by $b$, is set of indices where the words are not masked.
We remove those from $n_k$ elements if the intersection of $a$ and $b$ is empty.
With the remaining elements and $f$, we can approximately compute the value of $\beta$.


\begin{figure*}[ht]
\begin{subfigure}{.49\textwidth}
  \centering
  \includegraphics[width=0.95\linewidth]{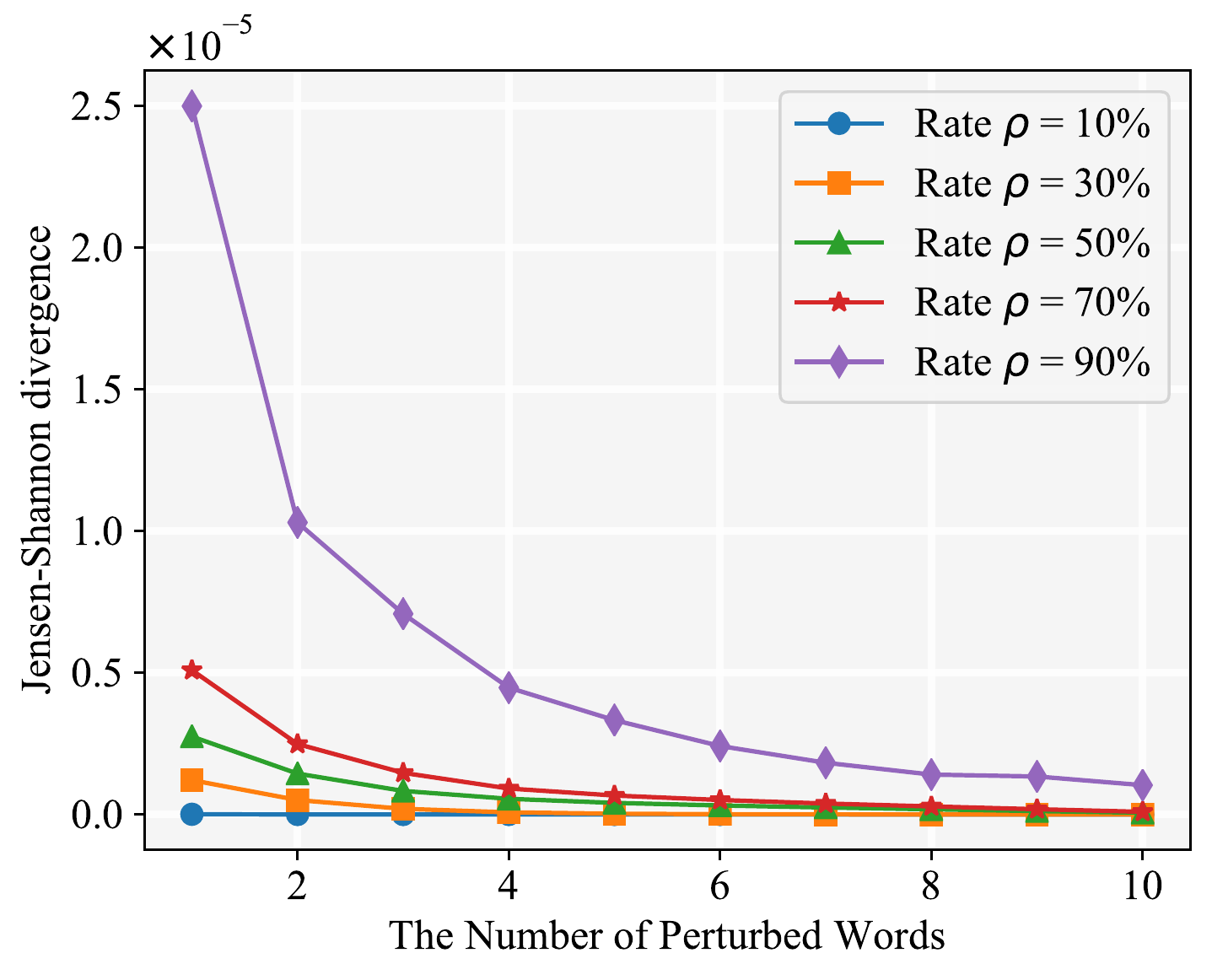}  
  \caption{AGNEWS}
  \label{fig:sub-agnews}
\end{subfigure}
\begin{subfigure}{.49\textwidth}
  \centering
  \includegraphics[width=0.95\linewidth]{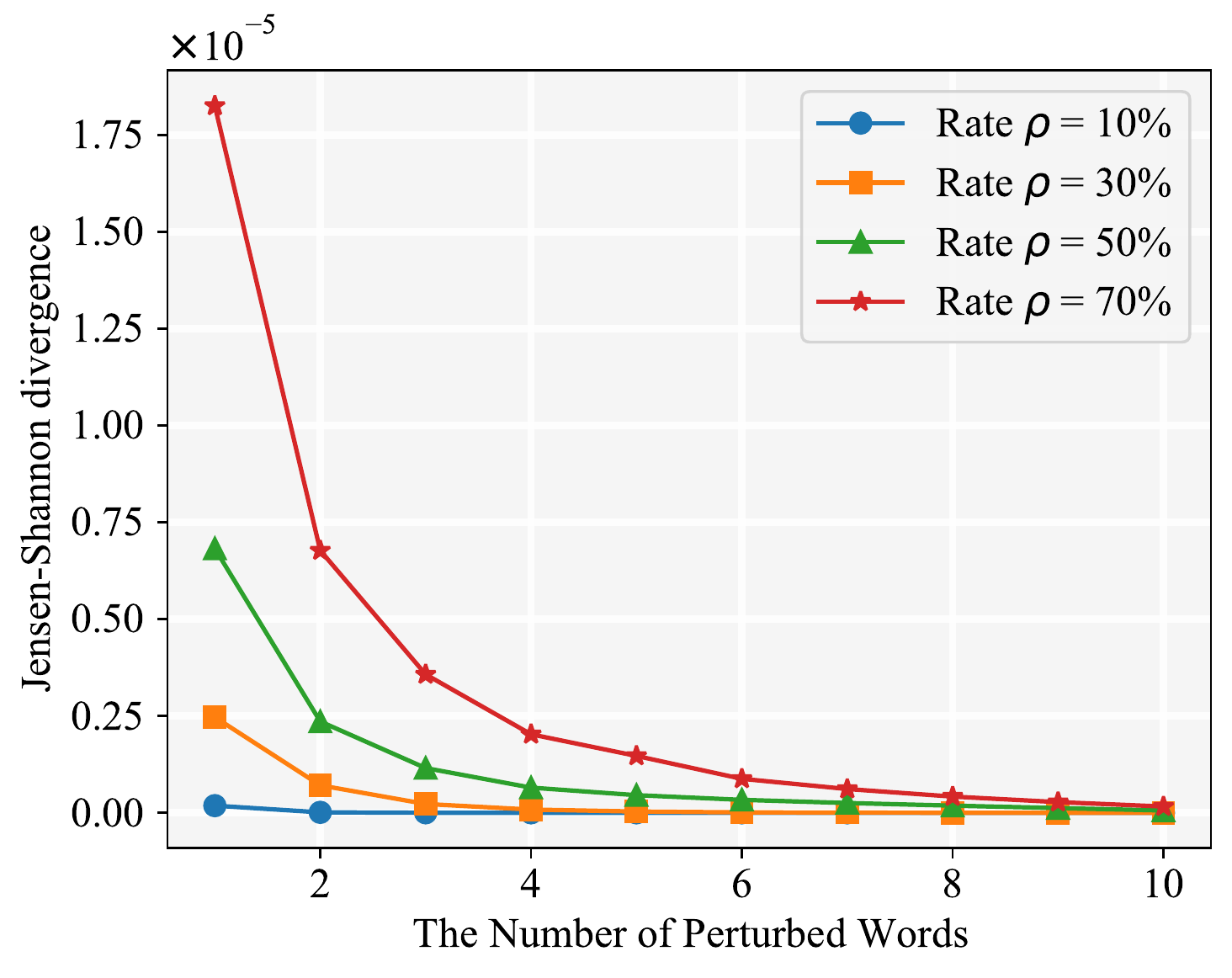}  
  \caption{SST2}
  \label{fig:sub-sst2}
\end{subfigure}
\caption{The Jensen$\text{-}$Shannon divergence between the values of $\beta$ and $p_c(\boldsymbol{x})$ estimated on two datasets (AGNEWS and SST2) with different masking rate $\rho$. No matter what value of $\rho$ is, all the divergence values are less than $2.5\times10^{-5}$ on AGNEWS dataset and $1.75\times10^{-5}$ on SST2 dataset. Therefore, we can use $p_c(\boldsymbol{x})$ to approximate the value of $\beta$ when the number of perturbed words is large enough. Note that we do not show the results when Rate $\rho = 90\%$ on SST2 here, because it is hard, even impossible to train and obtain the base classifier when $\rho \geq 90\%$ (see Section \ref{sec-certified-result} for discussion). }
\label{fig:js-divergence}
\end{figure*}


As the value of $r$ grows, for any $a$ it is more likely that $a$ is overlapped with any sampled $b$, and the value $\beta$ will approach to $p_c(\boldsymbol{x})$.
To investigate how close the values of $\beta$ and $p_c(\boldsymbol{x})$ are to each other, we conducted an experiment on the test set of both AGNEWS and SST2 datasets, in which $n_r = 200$ and $n_k = 10,000$, and use the Jensen-Shannon divergence
to calculate the distance of these two distributions.
As we can see from Figure \ref{fig:js-divergence}, no matter what value of $\rho$ is, all the Jensen-Shannon divergences values are very small and less than $2.5\times10^{-5}$ on AGNEWS and $1.75\times10^{-5}$ on SST2 when the number of perturbed words is large enough.
Therefore, we can use $p_c(\boldsymbol{x})$ to approximate the value of $\beta$, namely $\beta \approx p_c (\boldsymbol{x})$,
in all the following experiments.



\begin{algorithm*}[h]
\caption{For prediction and certification}
\label{alg_overall_conclusion}
\begin{algorithmic}[1]
\Procedure {ClassifierG} {{$\boldsymbol{x}, h_{\boldsymbol{x}}, k_{\boldsymbol{x}},f, n$}}
  \State $\mathcal{B} \gets$ sampling $n$ elements from $\mathcal{U}(h_{\boldsymbol{x}},k_{\boldsymbol{x}})$
  \State $counts \gets 0$ for each label $c \in \mathcal{Y}$
  \For { each $\mathcal{H}$ in $\mathcal{B}$ }
    \State $\boldsymbol{x}_{\text{mask}} \gets \mathcal{M}(\boldsymbol{x}, \mathcal{H})$, $c \gets f(\boldsymbol{x}_{\text{mask}})$
    \State $counts[c] \gets counts[c] + 1$
  \EndFor
  \State \Return $count$
\EndProcedure

\Procedure {Predict} {{$\boldsymbol{x}, h_{\boldsymbol{x}}, k_{\boldsymbol{x}},f, n$}}
  \State $counts \gets$ $\textsc{ClassifierG}(\boldsymbol{x}, h_{\boldsymbol{x}}, k_{\boldsymbol{x}},f, n)$
  \State $\hat{c} \gets $ the top index with the maximum $counts$
  \State $p_{\hat{c}} \gets counts[\hat{c}] / n$
  \State \Return $\hat{c}, p_{\hat{c}} $
\EndProcedure

\Procedure {Certify} {{$\boldsymbol{x},y, h_{\boldsymbol{x}}, k_{\boldsymbol{x}},f, n, n', \alpha$}}
  \State $\hat{c}, p_{\hat{c}} \gets \textsc{Predict}(\boldsymbol{x}, h_{\boldsymbol{x}}, k_{\boldsymbol{x}},f, n)$
  \If {$\hat{c} \neq y $} \Return ``N/A''
  \Else
  \State $counts \gets \textsc{ClassifierG}(\boldsymbol{x}, h_{\boldsymbol{x}}, k_{\boldsymbol{x}},f, n')$
  \State $\underline{p_y} \gets $ Using Eq. (\ref{equ-lower-bound}) with $counts[y], n'$ and $\alpha$
  \State $\beta \gets counts[y] / n'$
  \EndIf
  \For {$d$ from $0$ to $h_{\boldsymbol{x}}$}
    \State $\Delta \gets $ Using Eq. (\ref{equ-beta-definition}) with $h_{\boldsymbol{x}}, k_{\boldsymbol{x}}$ and $d$
    \If {$\underline{p_y} - \beta \Delta > 0.5$} $d \gets d + 1$
    \Else \hspace{0.5em} \textbf{break}
    \EndIf
  \EndFor
  \State \Return $d$
\EndProcedure
\end{algorithmic}
\end{algorithm*}

\subsection{Practical Algorithms}

In order for $g$ to classify the labeled examples correctly and robustly, the base classifier $f$ is trained to classify texts in which $\rho$ words are masked.
Specifically, at each training iteration, we first sample a mini-batch of samples and randomly performing mask operation on the samples.
We then apply gradient descent on $f$ based on the masked mini-batch.

We present practical Monte Carlo algorithms for evaluating $g(\boldsymbol{x})$ and certifying the robustness of $g$ around $\boldsymbol{x}$ in Algorithm \ref{alg_overall_conclusion}.
Evaluating the smoothed classifier's prediction $g(\boldsymbol{x})$ requires to identify the class $c$ with maximal weight in the categorical distribution.
The procedure described in pseudocode as $\textsc{Predict}$ that randomly draws $n$ masked copies of $\boldsymbol{x}$ and runs these $n$ copies through $f$.
If $c$ appeared more than any other class, then $\textsc{Predict}$ returns $c$.

 

Evaluating and certifying the robustness of $g$ around an input $\boldsymbol{x}$ requires not only identify the class $c$ with maximal weight, but also estimating a lower bound $\underline{p_c(\boldsymbol{x})}$ and $\beta$. 
In the procedure described as $\textsc{Certify}$, we first ensure that $g$ correctly classifies $\boldsymbol{x}$ as $y$, and then estimate the values of $\underline{p_c(\boldsymbol{x})}$ and $\beta$ by randomly generating $n'$ masked copies of $\boldsymbol{x}$, where $n'$ is much greater than $n$.
We gradually increase the value of $d$ (the number words can be perturbed) from $0$ by $1$, and compute $\Delta$ by Equation (\ref{equ-beta-definition}). 
This process will continue until $\underline{p_c(\boldsymbol{x})} - \beta \Delta \leq 0.5$ (see Corollary \ref{corollary-certified}), and when it stops $\textsc{Certify}$ returns $d$ as the maximum 
\emph{certified robustness} for $\boldsymbol{x}$.
In this way, we can certify with ($1 - \alpha$) confidence that $g(\boldsymbol{x}')$ will return the label $y$ for any adversarial example $\boldsymbol{x}'$ if $\norm{\boldsymbol{x} - \boldsymbol{x}'}_0 \leq d$.

\section{Experiments}
We first give the certified robustness of our RanMASK on AGNEWS \cite{zhang2015character} and SST2 \cite{socher2013recursive} datasets, and then report the empirical robustness on these datasets by comparing with other representative defense methods, including PGD-K \cite{madry2018towards}, FreeLB \cite{zhu2019freelb}, Adv-Hotflip \cite{ebrahimi-etal-2018-hotflip}, and adversarial data augmentation.
We conducted experiments on four datasets, AGNEWS \cite{zhang2015character} for text classification, SST2 \cite{socher2013recursive} for sentiment analysis, IMDB \cite{maas2011learning} for Internet movie review, and SNLI \cite{bowman-etal-2015-large} for natural language inference.
Finally, we empirically compare RanMASK with SAFER \cite{ye-etal-2020-safer}, a recently proposed certified defense that also can be applied to large architectures, such as BERT \cite{devlin2018bert}. 
We found that different randomized smoothing methods may behave quite differently when different ensemble methods are used.
We show that the ``majority-vote'' ensemble sometimes could fool the score-based attack algorithms in which the greedy search strategy is adopted. 
The improvement in the empirical robustness comes from not only the defense methods themselves, but also the type of ensemble method they use.

\subsection{Implementation Details}

Due to our randomized masking strategy is consistent with the large-scale pretrained masked language model used in BERT \cite{devlin2018bert}, we use BERT-based models, including BERT and RoBERTa \cite{liu2019roberta}, as our base models, which help to keep the performance of the base classifier $f$ to an acceptable level when it takes the masked texts as inputs since BERT-based models have the capacity to implicitly recover information about the masked words.

Unless otherwise specified, all the models are trained with AdamW optimizer \cite{loshchilov2017decoupled} with a weight decay $(1e-6)$, a batch size $32$, a maximum number of epochs $20$, and a gradient clip $(-1, 1)$ and a learning rate $(5e-5)$, which is decayed by the cosine annealing method \cite{loshchilov2016sgdr}.
All the models tuned on the validation set were used for testing and certifying.
We randomly selected $1,000$ test examples for each dataset in both certified and empirical experiments.
When conducting the experiments of certified robustness, we set the uncertainty $\alpha$ to $0.05$, the number of samples $n$ for $\textsc{Predict}$ procedure to $1,000$, the number of samples $n'$ for $\textsc{Certify}$ to $5,000$. 
To evaluate the empirical robustness of models, we set $n$ to $100$ to speed up the evaluation process.

\subsection{Results of the Certified Robustness}
\label{sec-certified-result}

We here provide the certified robustness of RanMASK on AGNEWS and SST2.
When reporting certified results, we refer to the following metrics introduced by Levine and Feizi \cite{levine2020robustness}:

\begin{itemize}
\normalsize
\item The \emph{certified robustness} of a text $\boldsymbol{x}$ is the maximum $d$ for which we can certify that the smoothed classifier $g(\boldsymbol{x}')$ will return the \emph{correct} label $y$ where $\boldsymbol{x}'$ is any adversarial example of $\boldsymbol{x}$ such that $\norm{\boldsymbol{x} - \boldsymbol{x}'}_0 \leq d$. 
If $g(\boldsymbol{x})$ labels $\boldsymbol{x}$ incorrectly, we define the certified robustness as ``N/A'' (see Algorithm \ref{alg_overall_conclusion}). 

\item The \emph{certified rate} of a text $\boldsymbol{x}$ is the \emph{certified robustness} of $\boldsymbol{x}$ divided by $\boldsymbol{x}$'s length $h_{\boldsymbol{x}}$.

\item The \emph{median certified robustness} (MCB) on a dataset is the median value of the \emph{certified robustness} across the dataset. 
It is the maximum $d$ for which the smoothed classifier $g$ can guarantee robustness for at least $50\%$ texts in the dataset. 
In other words, we can certify the classifications of over $50\%$ texts to be robust to any perturbation of at most $d$ words. 
When computing this median, the texts which $g$ misclassifies 
are counted as having $-\infinity$ certified robustness. For example, if the certified robustness of the texts in a dataset are $\{\text{N/A},\text{N/A},1,2,3\}$, the \emph{median certified robustness} is $1$, not $2$.

\item The \emph{median certified rate} (MCR) on a dataset is the median value of the \emph{certified rate} across the datasets, which is obtained in a similar way to MCB.

\end{itemize}

We first tested the robust classification on AGNEWS using RoBERTa \cite{liu2019roberta} as the base classifier. 
As we can see from Table \ref{tb-agnews-certified}, the maximum MCB was achieved at $5$ when using the masking rate $\rho = 90\%$ or $\rho = 95\%$, indicating that we can certify the classifications of over $50\%$ sentences to be robust to any perturbation of at most $5$ words. 
We chose to use the model ($\rho = 90\%$) to evaluate the empirical robustness on AGNEWS because it gives better classification accuracy on the clean data.


\begin{table}[htbp]
\small
\setlength{\abovecaptionskip}{0.05cm}
\begin{center}
\setlength{\tabcolsep}{0.7mm}
\begin{tabular}{c|>{\centering\arraybackslash}p{1.8cm}|>{\centering\arraybackslash}p{1.2cm}|>{\centering\arraybackslash}p{1.8cm}}
\hline
\hline
{\bf Rate $\rho$\%} & {\bf Accuracy\%} & {\bf MCB} & {\bf MCR\%}
\\ \hline
$40$ & $96.2$  & $1$ & $2.6$ \\
$50$ & $95.7$  & $1$ & $2.7$ \\
$60$ & $95.7$  & $2$ & $5.0$ \\
$65$ & $95.0$  & $2$ & $5.0$ \\
$70$ & $94.5$  & $2$ & $5.0$ \\
$75$ & $93.9$  & $3$ & $7.0$ \\
$80$ & $92.0$  & $3$ & $7.5$ \\
$85$ & $92.2$  & $4$ & $8.8$ \\
$\bf 90$ & $\bf 91.1$  & $\bf 5$ & $\bf 11.4$ \\
$95$ & $85.8$  & $5$ & $11.8$ \\
\hline
\hline
\end{tabular}
\end{center}
\caption{Robustness certificates on AGNEWS with different masking rates $\rho$. 
The maximum median certified robustness (MCB) was achieved at $5$ words when using $\rho = 90\%$ or $95\%$.
We use the model (highlighted in bold) when evaluating the empirical robustness against adversarial attacks.
``MCR'' denotes the median certified rate.
}
\label{tb-agnews-certified} 
\end{table}

We evaluated the robust classification on SST2 using RoBERTa as the base classifier too. 
As shown in Table \ref{tb-sst2-certified}, the maximum MCB was achieved at $2$ when $\rho = 70\%$ or $80\%$, indicating that over $50\%$ sentences are robust to any perturbation of $2$ words.
But, these two models achieve the maximum MCB in a higher cost of clean accuracy (about $10\%$ drop compared with the best).
We chose to use the model ($\rho = 30\%$) to evaluate the empirical robustness on SST2 due to its higher classification accuracy on the clean data.
We found that it is impossible to train the models when $\rho \geq 90\%$.
Unlike AGNEWS (created for the news topic classification), SST2 was constructed for the sentiment analysis.
The sentiment of a text largely depends on whether few specific sentiment words occur in the text.
All the sentiment words in a text would be masked with high probability when a large masking rate is applied (say $\rho \geq 90\%$), which makes it hard for any model to correctly predict the sentiment of the masked texts.



\begin{table}[htbp]
\small
\setlength{\abovecaptionskip}{0.05cm}
\begin{center}
\setlength{\tabcolsep}{0.7mm}
\begin{tabular}{c|>{\centering\arraybackslash}p{1.8cm}|>{\centering\arraybackslash}p{1.2cm}|>{\centering\arraybackslash}p{1.8cm}}
\hline
\hline
{\bf Rate $\rho$\%} & {\bf Accuracy\%} & {\bf MCB} & {\bf MCR\%}
\\
\hline
$20$ & $92.4$  & $1$ & $5.26$ \\
$\bf 30$ & $\bf 92.4$  & $\bf 1$ & $\bf 5.26$ \\
$40$ & $91.2$  & $1$ & $5.26$ \\
$50$ & $89.3$  & $1$ & $5.56$ \\
$60$ & $84.3$  & $1$ & $7.41$ \\
$70$ & $83.3$  & $2$ & $8.00$ \\
$80$ & $81.4$  & $2$ & $10.00$ \\
$90$ & $49.6$  & N/A & N/A \\
\hline
\hline
\end{tabular}
\end{center}
\caption{Robustness certificates on SST2 with different masking rates $\rho$.
The maximum median certified robustness (MCB) was achieved at $2$ words when $\rho = 70\%$ or $80\%$. 
We use the model (highlighted in bold) when evaluating the empirical robustness against adversarial attacks because $\rho = 30\%$ gives the better classification. ``MCR'' denotes the median certified rate.}
\label{tb-sst2-certified} 
\end{table}


\subsection{Results of the Empirical Robustness}
\label{sec:empirical_result}

As mentioned in the introduction, if a relatively small proportion of words are perturbed (relative to the masking rate $\rho$), then it is highly unlikely that all of the perturbed words can survive from the random masking operation.
As the value of $\rho$ decreases, we have to make sure that the following \emph{risk probability} is greater than $0.5$ for any text $\boldsymbol{x}$; otherwise more than $50\%$ masked copies of $\boldsymbol{x}$ will contain all the perturbed words, which easily causes the base classifier $f$ to make mistakes, and so does $g$.
\begin{equation}
\label{eq:probability}
\mathbb{P}(\boldsymbol{x}, \rho, \gamma) = \frac{
\tbinom {(1 - \gamma) h_{\boldsymbol{x}}} {(1 - \gamma - \rho) h_{ \boldsymbol{x}}}
} 
{ \tbinom{h_{ \boldsymbol{x}}}{\rho h_{ \boldsymbol{x}}}}
\end{equation}
\noindent where $\gamma$ is the maximum percentage of words that can be perturbed.
For AGNEWS dataset, this risk probability is very close to zero when $\rho = 90\%$ (chosen in the certified robustness experiment) no matter what the value of $\gamma$ is applied. 
We use the average length of texts in a dataset (instead of each text) to estimate the risk probability.
For SST2, this risk probability is approximately equal to $0.5$ when $\rho = 30\%$ (selected in the certified robustness experiment).
To reduce this risk, we designed a new sampling strategy in which the probability of a word being masked corresponds to its output probability of a BERT-based language model (LM).
Generally, the higher the output probability of a word is, the lower the probability that this word is expected to be perturbed.
We would like to retain the words that have not been perturbed as much as possible.
This LM-based sampling instead of that based on the uniform distribution was used to evaluate the empirical results on SST2. 
Note that LM-based sampling is unnecessary when evaluating on AGNEWS, and the experimental results also show that there is no significant difference whether or not the LM-based sampling is used on this dataset.

In the following experiments, we consider two ensemble methods \cite{cheng2020voting}: \emph{logits-summed} ensemble (logit) and \emph{majority-vote} ensemble (vote). 
In the ``logit'' method, we take the average of the logits produced by the base classifier $f$ over all the individual random samples as the final prediction. 
In the ``vote'' strategy, we simply count the votes for each class label.
The following metrics are used to report the empirical results:
\begin{itemize}
\normalsize
\item The \emph{clean accuracy} (Cln) is the classification accuracy achieved on the clean texts $\boldsymbol{x}$.
\item The \emph{robust accuracy} (Boa) is the accuracy of a classifier achieved under a certain attack.
\item The \emph{success rate} (Succ) is the number of texts successfully perturbed by an attack algorithm (causing the model to make errors) divided by all the number of texts to be attempted.
\end{itemize}

We evaluate the empirical robustness under test-time attacks by using TextAttack\footnote{\url{https://github.com/QData/TextAttack}} 
framework \cite{morris2020textattack} 
with three black-box, score-based attack algorithms: TextFooler \cite{Jin_Jin_Zhou_Szolovits_2020}, BERT-Attack \cite{li2020bert}, and DeepWordBug \cite{gao2018black}. 
TextFooler and BERT-Attack adversarially perturb the text inputs by the word-level substitutions, whereas DeepWordBug performs the character-level perturbations to the inputs.
TextFooler generates synonyms using $50$ nearest neighbors of GloVe vectors \cite{pennington2014glove}, while
BERT-Attack uses BERT to generate synonyms dynamically, meaning that no defenders can know the synonyms used by BERT-Attack in advance.
DeepWordBug generates text adversarial examples by replacing, scrambling, and erasing few characters of some words in the input texts.

We compare RanMASK with the following four defense methods proposed recently:

\begin{table*}[htbp]
\small
\setlength{\abovecaptionskip}{0.05cm}
\begin{center}
\setlength{\tabcolsep}{0.7mm}
\begin{tabular}{l|*{8}{>{\centering\arraybackslash}p{1.0cm}|}{>{\centering\arraybackslash}p{1.0cm}}}
\hline
\hline
\multirow{2}{*}{\bf Method} & \multicolumn{3}{c|}{\bf TextFooler} & \multicolumn{3}{c|}{\bf BERT-Attack} &\multicolumn{3}{c}{\bf DeepWordBug} \\
\cline{2-10}
  &\bf Cln\%   &\bf Boa\%  &\bf Succ\%  &\bf Cln\%   &\bf Boa\%  &\bf Succ\%  &\bf Cln\%   &\bf Boa\%  &\bf Succ\%  \\
\hline
Baseline (RoBERTa) & $93.9$ & $15.8$ & $83.2$ & $94.7$ & $26.7$ & $71.8$ & $94.2$ & $33.0$ & $65.0$ \\
PGD-10 \cite{madry2018towards} & $\bf 95.0$ & $22.3$ & $76.5$ & $\bf 95.3$ & $30.0$ & $68.5$ & $\bf 94.9$ & $38.8$ & $59.1$ \\
FreeLB \cite{zhu2019freelb} & $93.9$ & $24.6$ & $73.8$ & $\bf 95.3$ & $28.3$ & $70.3$ & $93.7$ & $44.0$ & $53.0$ \\
Adv-Hotflip \cite{ebrahimi-etal-2018-hotflip} & $93.4$ & $21.3$ & $77.2$ & $93.9$ & $26.8$ & $71.5$ & $94.6$ & $37.6$ & $60.3$ \\
Data Augmentation & $93.3$ & $23.7$ & $74.6$ & $92.3$ & $39.1$ & $57.6$ & $93.8$ & $49.7$ & $47.0$ \\
RanMASK-$90\%$ (logit) & $89.1$ & $42.7$ & $52.1$ & $88.5$ & $30.0$ & $66.1$ & $89.8$ & $45.4$ & $45.4$ \\
RanMASK-$90\%$ (vote) & $91.2$ & $\bf 55.1$ & $\bf 39.6$ & $89.1$ & $\bf 41.1$ & $\bf 53.9$ & $90.3$ & $\bf 57.5$ & $\bf 36.0$ \\
\hline
\hline
\end{tabular}
\end{center}
\caption{Empirical results on AGNEWS. RanMASK-$90 \%$ with the ``vote'' ensemble method achieved the best results on the robust accuracy under all three different attack algorithms, indicating that RanMASK can defend against both word substitution-based attacks and character-level perturbations.
}
\label{tb-agnews-emperical} 
\end{table*}

\begin{table*}[htbp]
\small
\setlength{\abovecaptionskip}{0.05cm}
\begin{center}
\setlength{\tabcolsep}{0.7mm}
\begin{tabular}{l|*{8}{>{\centering\arraybackslash}p{1.0cm}|}{>{\centering\arraybackslash}p{1.0cm}}}
\hline
\hline
\multirow{2}{*}{\bf Method} & \multicolumn{3}{c|}{\bf TextFooler} & \multicolumn{3}{c|}{\bf BERT-Attack} &\multicolumn{3}{c}{\bf DeepWordBug} \\
\cline{2-10}
  &\bf Cln\%   &\bf Boa\%  &\bf Succ\%  &\bf Cln\%   &\bf Boa\%  &\bf Succ\%  &\bf Cln\%   &\bf Boa\%  &\bf Succ\%  \\
\hline
Baseline (RoBERTa) & $\bf 94.3$ & $5.4$ & $94.3$ & $93.9$ & $6.2$ & $93.4$ & $\bf 94.7$ & $17.0$ & $82.1$ \\
PGD-10 \cite{madry2018towards} & $94.0$ & $5.6$ & $94.0$ & $\bf 94.4$ & $5.6$ & $94.1$ & $92.9$ & $18.3$ & $80.3$ \\
FreeLB \cite{zhu2019freelb} & $93.7$ & $13.9$ & $85.2$ & $93.8$ & $10.4$ & $89.0$ & $93.0$ & $23.7$ & $74.5$ \\
Adv-Hotflip \cite{ebrahimi-etal-2018-hotflip} & $\bf 94.3$ & $12.3$ & $87.0$ & $93.8$ & $11.4$ & $87.9$ & $93.3$ & $23.4$ & $74.9$ \\
Data Augmentation & $91.0$ & $9.6$ & $89.5$ & $88.2$ & $16.9$ & $80.8$ & $91.8$ & $23.5$ & $74.4$ \\
RanMASK-$30\%$ (logit) & $92.9$ & $8.9$ & $90.4$ & $92.9$ & $9.5$ & $89.8$ & $93.0$ & $21.1$ & $77.3$ \\
RanMASK-$30\%$ (vote) & $92.7$ & $12.9$ & $86.1$ & $93.0$ & $11.4$ & $87.7$ & $92.7$ & $27.5$ & $70.3$ \\
RanMASK-$30\%$ (vote) + LM & $90.6$ & $\bf 23.4$ & $\bf 74.2$ & $90.4$ & $\bf 22.8$ & $\bf 74.8$ & $91.0$ & $\bf 41.7$ & $\bf 53.1$ \\
\hline
\hline
\end{tabular}
\end{center}
\caption{Empirical results on SST2. RanMASK-$30 \%$ with the ``vote'' ensemble method and the LM-based sampling strategy achieved the best results on the robust accuracy under all three different attack algorithms. 
}
\label{tb-sst2-emperical} 
\end{table*}

\begin{itemize}
\setlength{\itemsep}{0pt}
\setlength{\parsep}{0pt}
\setlength{\parskip}{0pt}
\item PGD-K \cite{madry2018towards}: applies a gradient-guided adversarial perturbations to word embeddings and minimizes the resultant adversarial loss inside different regions around input samples. 

\item FreeLB \cite{zhu2019freelb}: adds norm-bounded adversarial perturbations to the input's word embeddings using a gradient-based method, and enlarges the batch size with diversified adversarial samples under such norm constraints.

\item Adv-Hotflip \cite{ebrahimi-etal-2018-hotflip}: first generates textual adversarial examples by using Hotflip \cite{ebrahimi2017hotflip} and then augments the generated examples with the original training data to train a robust model.
Unlike PGD-K and FreeLB, Adv-Hotflip will generate real adversarial examples by replacing the original words with their synonyms rather than performing adversarial perturbations in the word embedding space.

\item Adversarial Data Augmentation: it still is one of the most successful defense methods for NLP models~\cite{miyato2016adversarial, sato2018interpretable}.
During the training time, they replace a word with one of its synonyms that maximizes the prediction loss.
By augmenting these adversarial examples with the original training data, the model is robust to such perturbations.
\end{itemize}


The results of the empirical robustness on AGNEWS listed shown in Table \ref{tb-agnews-emperical}. 
RanMASK-$90\%$ consistently performs better than the competitors under all the three attack algorithms on the robust accuracy while suffering little performance drop on the clean data. 
The empirical results on SST2 are reported in Table \ref{tb-sst2-emperical}, and we found similar trends as those on AGNEWS, especially for those when the LM-based sampling was applied.
On IMDB, we even observed that RanMASK achieved the highest accuracy on the clean data with $1.5\%$ improvement comparing to the baseline built upon RoBERTa, where the masking rate (i.e., $30\%$) was tuned on the validation set when the maximum MCB was achieved as the method introduced in Section \ref{sec-certified-result}.
The results for three text classification datasets show that our RanMASK consistently achieves better robust accuracy while suffering little loss on the original clean data.

Comparing to the baseline, RanMASK can improve the accuracy under attack or the robust accuracy (Boa) by $21.06\%$, and lower the attack success rate (Succ) by $23.71\%$ on average at the cost of $2.07\%$ decrease in the clean accuracy across three datasets and under three attack algorithms.
When comparing to a strong competitor, FreeLB \cite{zhu2019freelb} that was proposed very recently, RanMASK still in overall can increase the accuracy under attack by $15.47\%$, and reduce the attack success rate by $17.71\%$ on average at the cost of $1.98\%$ decrease in the clean accuracy.
Generally, RanMASK with the ``vote'' ensemble performs better than that with the ``logit'' ensemble, except on IMDB dataset under BERT-Attack and TextFooler attacks. 
We will thoroughly discuss the properties and behaviors of those two ensemble methods in the following two sections.

\begin{table*}[htbp]
\small
\setlength{\abovecaptionskip}{0.05cm}
\begin{center}
\setlength{\tabcolsep}{0.7mm}
\begin{tabular}{l|*{8}{>{\centering\arraybackslash}p{1.0cm}|}{>{\centering\arraybackslash}p{1.0cm}}}
\hline
\hline
\multirow{2}{*}{\bf Method} & \multicolumn{3}{c|}{\bf TextFooler} & \multicolumn{3}{c|}{\bf BERT-Attack} &\multicolumn{3}{c}{\bf DeepWordBug} \\
\cline{2-10}
  &\bf Cln\%   &\bf Boa\%  &\bf Succ\%  &\bf Cln\%   &\bf Boa\%  &\bf Succ\%  &\bf Cln\%   &\bf Boa\%  &\bf Succ\%  \\
\hline
Baseline (RoBERTa) & $91.5$ & $0.5$ & $99.4$ & $91.5$ & $0.0$ & $100.0$ & $91.5$ & $48.5$ & $47.0$ \\
PGD-10 \cite{madry2018towards} & $92.0$ & $1.0$ & $98.9$ & $92.0$ & $0.5$ & $99.4$ & $92.0$ & $44.5$ & $51.6$ \\
FreeLB \cite{zhu2019freelb} & $92.0$ & $3.5$ & $96.2$ & $92.0$ & $2.5$ & $97.3$ & $92.0$ & $52.5$ & $42.9$ \\
Adv-Hotflip \cite{ebrahimi-etal-2018-hotflip} & $91.5$ & $6.5$ & $92.9$ & $91.5$ & $11.5$ & $87.4$ & $91.5$ & $42.5$ & $53.5$ \\
Data Augmentation & $90.5$ & $2.5$ & $97.2$ & $91.0$ & $5.5$ & $94.0$ & $91.0$ & $50.5$ & $44.5$ \\
RanMASK-$30\%$ (logit) & $\bf 93.0$ & $\bf 23.5$ & $\bf 74.
7$ & $93.0$ & $\bf 22.0$ & $\bf 76.3$ & $\bf 93.5$ & $62.0$ & $33.7$ \\
RanMASK-$30\%$ (vote) & $\bf 93.0$ & $18.0$ & $80.7$ & $\bf 93.5$ & $17.0$ & $81.8$ & $92.5$ & $\bf 66.0$ & $\bf 28.7$ \\
\hline
\hline
\end{tabular}
\end{center}
\caption{Empirical results on IMDB. RanMASK-$30 \%$ with the ``vote'' ensemble method achieved the best results on the robust accuracy under all three different attack algorithms. 
}
\label{tb-imdb-emperical} 
\end{table*}

\begin{table*}[htbp]
\small
\setlength{\abovecaptionskip}{0.05cm}
\begin{center}
\setlength{\tabcolsep}{0.7mm}
\begin{tabular}{l|*{8}{>{\centering\arraybackslash}p{1.0cm}|}{>{\centering\arraybackslash}p{1.0cm}}}
\hline
\hline
\multirow{2}{*}{\bf Method} & \multicolumn{3}{c|}{\bf TextFooler} & \multicolumn{3}{c|}{\bf BERT-Attack} &\multicolumn{3}{c}{\bf DeepWordBug} \\
\cline{2-10}
  &\bf Cln\%   &\bf Boa\%  &\bf Succ\%  &\bf Cln\%   &\bf Boa\%  &\bf Succ\%  &\bf Cln\%   &\bf Boa\%  &\bf Succ\%  \\
\hline
Baseline (RoBERTa) & $91.0$ & $3.9$ & $95.7$ & $91.0$ & $0.6$ & $99.3$ & $91.0$ & $4.2$ & $95.4$ \\
PGD-10 \cite{madry2018towards} & $\bf 91.9$ & $4.7$ & $95.0$ & $\bf 91.9$ & $1.0$ & $98.9$ & $\bf 91.9$ & $4.8$ & $94.8$ \\
FreeLB \cite{zhu2019freelb} & $91.2$ & $4.3$ & $95.3$ & $91.2$ & $0.4$ & $99.6$ & $91.2$ & $5.3$ & $94.1$ \\
Adv-Hotflip \cite{ebrahimi-etal-2018-hotflip} & $88.8$ & $6.0$ & $93.2$ & $88.8$ & $1.5$ & $98.3$ & $88.5$ & $8.9$ & $90.0$ \\
Data Augmentation & $89.5$ & $14.2$ & $84.1$ & $89.7$ & $1.8$ & $98.0$ & $91.0$ & $20.3$ & $77.7$ \\
RanMASK-$15\%$ (logit) & $89.4$ & $10.8$ & $87.9$ & $89.8$ & $1.2$ & $98.7$ & $89.7$ & $6.8$ & $92.4$ \\
RanMASK-$15\%$ (vote) & $87.0$ & $\bf 21.5$ & $\bf 74.7$ & $89.5$ & $\bf 5.8$ & $\bf 93.5$ & $86.2$ & $\bf 23.0$ & $\bf 73.3$ \\
\hline
\hline
\end{tabular}
\end{center}
\caption{Empirical results on SNLI. RanMASK-$15 \%$ with the ``vote'' ensemble method achieved the best results on the robust accuracy under all three different attack algorithms. 
}
\label{tb-snli-emperical} 
\end{table*}

Any model performed on IMDB shows to be more vulnerable to adversarial attacks than the same one on AGNEWS.
For example, BERT-Attack achieved $100\%$ attack success rate against the baseline model on IMDB while its attack success rates are far below $100\%$ on the other datasets.
It is probably because the average length of the sentences in IMDB ($255$ words on average) is much longer than that in AGNEWS ($43$ words on average).
Longer sentences allow the adversaries to apply more word substitution-based or character-level perturbations to the examples.
While RanMASK shows to be more resistant to adversarial attacks, it also can improve the clean accuracy on IMDB. 
One reasonable explanation is that the models rely too heavily on the non-robust features that are less relevant to the categories to be classified, and our random masking strategy disproportionately affects non-robust features which thus hinders the model’s reliance on them.
Note that the sentences in IMDB are relatively long, and many words in any sentence might be irrelevant for the classification, but would be used by the models for the prediction.

We also conducted the experiments of natural language inference on Stanford Natural Language Inference (SNLI) \cite{bowman2015large} corpus, which is a collection of $570,000$ English sentence pairs (a premise and a hypothesis) manually labeled for balanced classification with the labels entailment, contradiction, and neutral.
What makes the natural language inference different from text classification tasks is that it needs to determine whether the directional relation holds whenever the truth of one text (i.e., hypothesis) follows from another text (i.e., premise). 
We implemented a baseline model based on RoBERTa for this task.
The premise and hypothesis are encoded by running RoBERTa on the word embeddings to generate the sentence representations, which uses attention between the premise and hypothesis to compute richer representations of each word in both sentences, and then the concatenation of these encodings are fed to a two-layer feedforward network for the prediction.
The baseline model was trained with cross-entropy loss, and their hyperparameters were tuned on the validation set.

The results of the empirical robustness on SNLI are reported in Table \ref{tb-snli-emperical}. 
The masking rate (i.e., $15\%$) was tuned for RanMASK on the validation set when the maximum MCB was achieved. 
From these numbers, a handful of trends are readily apparent.
RanMASK using the ``vote'' ensemble achieved better empirical robustness than that using the ``logit'' ensemble again.
Comparing to the baseline, RanMASK can improve the accuracy under attack or the robust accuracy (Boa) by $14.05\%$, and lower the attack success rate (Succ) by $16.30\%$ on average at the cost of $3.43\%$ decrease in the clean accuracy under three different attack algorithms.
When FreeLB is used for comparison, RanMASK still can improve the robust accuracy (Boa) by $13.43\%$, and reduce the attack success rate (Succ) by $15.83\%$ on average.
In conclusion, RanMASK can significantly improve the accuracy under different attacks much further than existing defense methods on various tasks, including text classification, sentiment analysis, and natural language inference. 



\subsection{Comparison with SAFER}
\label{sec:safer}

\begin{table*}[t]
\small
\setlength{\abovecaptionskip}{0.05cm}
\begin{center}
\setlength{\tabcolsep}{0.7mm}
\begin{tabular}{l|*{8}{>{\centering\arraybackslash}p{1.0cm}|}{>{\centering\arraybackslash}p{1.0cm}}}
\hline
\hline
\multirow{2}{*}{\bf Method} & \multicolumn{3}{c|}{\bf TextFooler} & \multicolumn{3}{c|}{\bf \footnotesize BERT-Attack} & \multicolumn{3}{c}{\bf \footnotesize DeepWordBug} \\
\cline{2-10}
 &\bf Cln\%   &\bf Boa\%  &\bf Succ\% &\bf Cln\%   &\bf Boa\%  &\bf Succ\% &\bf Cln\%   &\bf Boa\%  &\bf Succ\% \\
\hline
Baseline (BERT) & $93.0$ & $5.6$ & $94.0$ & $95.1$ & $16.3$ & $82.9$ & $94.3$ & $16.6$ & $82.4$ \\
\hline
SAFER (logit) & $94.6$ & $26.1$ & $72.4$ & $94.8$ & $29.0$ & $69.4$ & $95.1$ & $31.9$ & $66.5$ \\
SAFER (vote) & $\bf 95.4$ & $\bf 78.6$ & $\bf 17.6$ & $94.3$ & $63.4$  & $32.8$ & $95.2$ & $78.4$  & $17.7$\\
\hline
RanMASK-$90\%$ (logit) & $91.3$ & $47.3$ & $48.2$ & $91.6$ & $38.3$ & $58.2$ & $89.2$ & $39.6$ & $55.6$  \\
RanMASK-$90\%$ (vote) & $90.3$ & $51.9$  & $42.5$ & $92.7$ & $46.3$ & $50.1$  & $90.4$ & $51.8$ & $42.7$  \\
\hline
RanMASK-$5\%$ (logit) & $94.4$ & $13.2$ & $86.0$ & $\bf 96.0$ & $25.6$ & $73.3$ & $94.8$ & $23.4$ & $75.3$ \\
RanMASK-$5\%$ (vote) & $93.9$ & $68.6$ & $26.9$ & $95.2$ & $63.0$ & $33.8$ & $\bf 95.3$ & $77.1$ & $19.1$ \\
RanMASK-$5\%$ (vote) + LM & $94.8$ & $71.4$ & $24.7$ & $95.7$ & $\bf 65.3$ & $\bf 31.8$ & $93.8$ & $\bf 80.3$ & $\bf 14.4$ \\
\hline
\hline
\end{tabular}
\end{center}
\caption{Empirical results of RanMASK on AGNEWS compared with SAFER under three attack algorithms: TextFooler, BERT-Attack and DeepWordBug. 
}
\label{tb-safer} 
\end{table*}

We report in Table \ref{tb-safer} the empirical robustness of RanMASK on AGNEWS compared with SAFER, a very recently proposed certified defense \cite{ye-etal-2020-safer}.
From these numbers, we found that RanMASK outperforms SAFER under the setting where the ``logit'' ensemble is used, while SAFER slightly performs better than RanMASK when using the ``vote'' ensemble for the predictions under the attack of TextFooler.
However, this comparison is not direct and fair.
First, SAFER makes use of the same synonym table used by TextFooler. 
Second, we found that different smoothing defense methods behave quite differently as the ensemble method is changed from the ``vote'' to the ``logit.''



Typical score-based attack algorithms, such as TextFooler and DeepWordBug, usually use two steps to craft adversarial examples: greedily identify the most vulnerable position to change; modify it slightly to maximize the model's prediction error. 
This two-step would be repeated iteratively until the model's prediction changes. 
If the ``majority-vote'' method is used, the class distributions produced by the models trained with SAFER would be quite sharp, even close to one-hot categorical distribution\footnote{In contrast to SAFER, the class distributions produced by the models trained with RanMASK are relatively smoother than those with SAFER.
We estimated the average entropy of the distributions predicted by SAFER and RanMASK on $1,000$ test samples selected randomly from AGNEWS.
When TextFooler starts to attack, the average entropy of SAFER's predictions is $0.006$, while those of RanMASK's are $0.025$, $0.036$, $0.102$, and $0.587$ when $\rho = 5\%$, $10\%$, $50\%$, and $90\%$ respectively.
Note that the greater the entropy is, the smoother the distribution will be.}, which hinders the adversaries to peek the changes in the model's predictions by a small perturbation on the input, ending up trapped in local minima. 
This forces the adversaries to launch the decision-based attacks \cite{maheshwary2020generating} instead of the score-based ones, which can dramatically affect the resulting attack success rates.
If the ``logit'' ensemble method is used or the attack algorithm is designed to perturb more than one word at a time, the empirical robustness of SAFER will drop significantly. Therefore, it is unfair to compare ``majority vote''-based ensemble defense methods with others when conducting empirical experiments. We believe these methods using the ``majority vote'' ensemble will greatly improve model's defense performance when the models are deployed for real-world applications, but we recommend using ``logit'' ensemble methods if one needs to analyze and prove the effectiveness of the proposed algorithm against textual adversarial attacks in future research.

\subsection{Impact of Masking Rate on Robust Accuracy}

\begin{figure*}[ht]
\begin{subfigure}{.49\textwidth}
  \centering
  \includegraphics[width=0.9\linewidth]{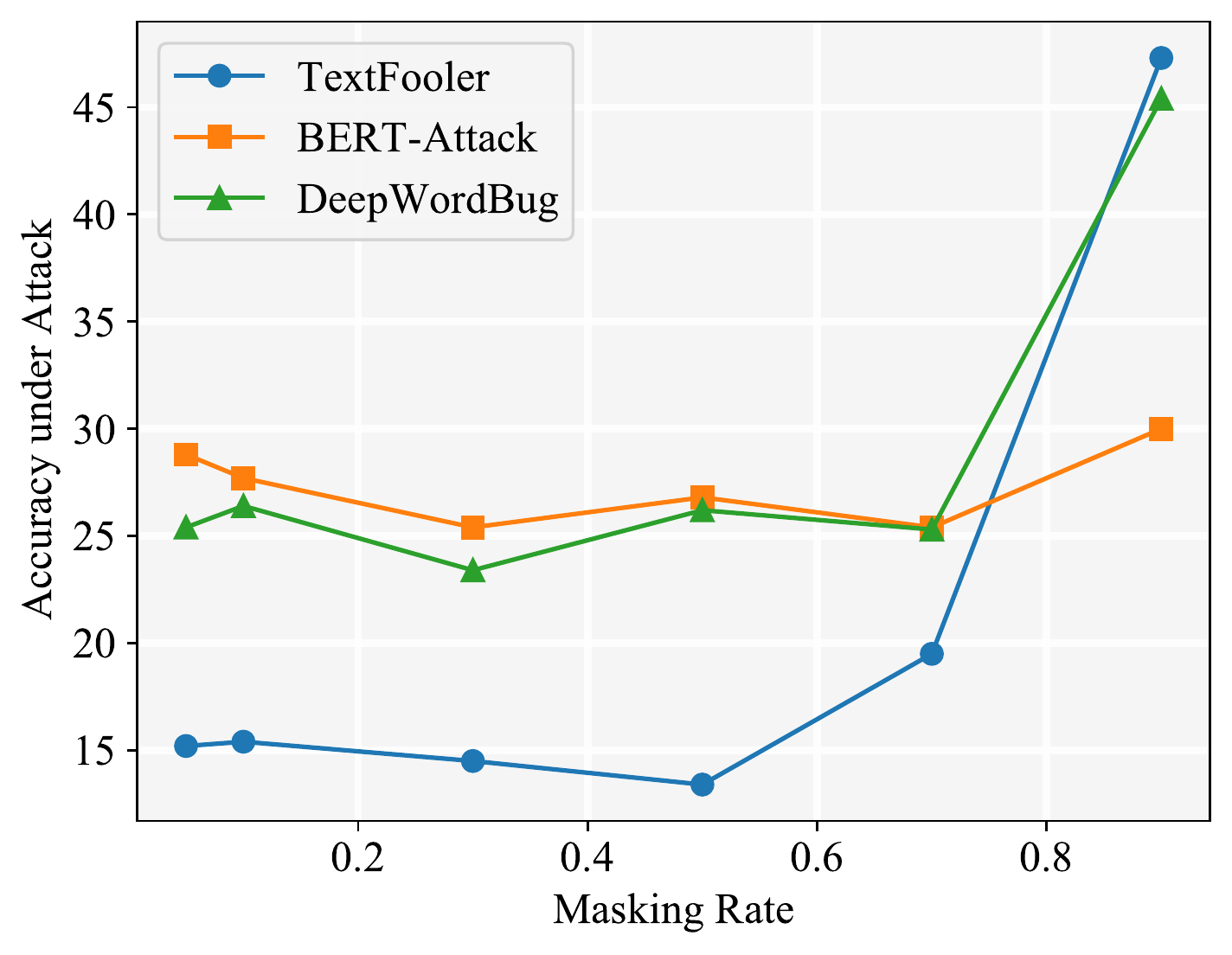}  
  \caption{Logits-summed Ensemble}
  \label{fig:sub-agnews-ensemble}
\end{subfigure}
\begin{subfigure}{.49\textwidth}
  \centering
  \includegraphics[width=0.9\linewidth]{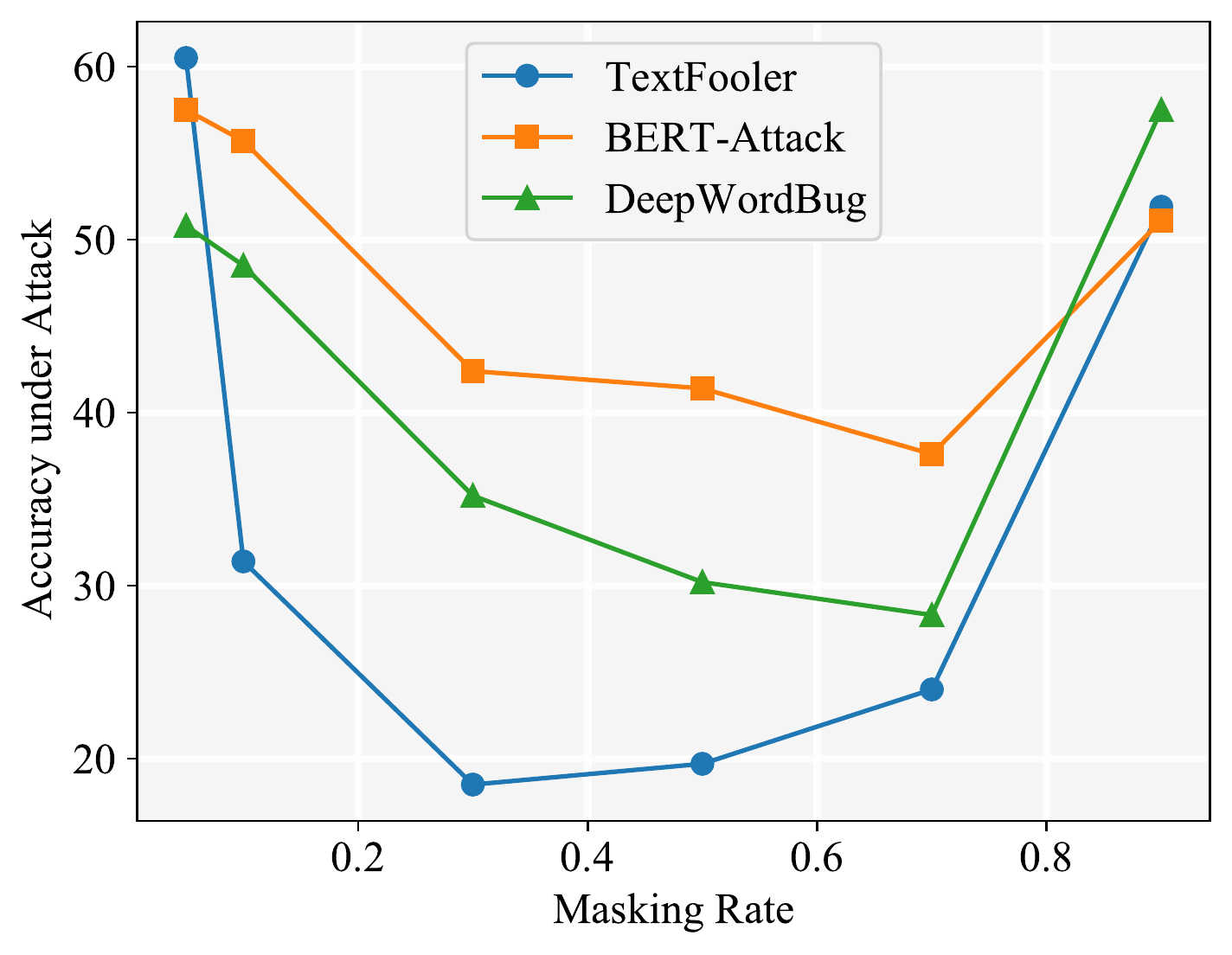}  
  \caption{Majority-vote Ensemble}
  \label{fig:sub-sst2-ensemble}
\end{subfigure}
\caption{Accuracy under attack versus masking rates}
\label{fig:rate_ablation}
\end{figure*}

We want to understand the impact of different masking rates on the accuracy of RanMASK under attack by varying the masking rates from $5\%$ to $90\%$.
We show in Figure \ref{fig:rate_ablation} the average robust accuracy of RanMASK on the test set of AGNEWS versus the masking rates with two ensemble methods under three different attacks: TextFooler, BERT-Attack, and DeepWordBug. 
The average accuracy is obtained over $5$ runs with different random initialization for each setting.
When the ``logit'' ensemble method is applied, the accuracy under attack generally increases until the best performance is achieved at the masking rate of ninety percent ($\rho = 90\%$) and there is a big jump from $70\%$ to $90\%$.
However, if the ``vote'' ensemble method is used, we observed a dramatic decrease in the robust accuracy when the masking rates vary from $5\%$ to $30\%$, and then the accuracy climbs up smoothly until getting to its peak when $\rho = 90\%$.
Although it seems counterintuitive that the robust accuracy falls greatly at first and rises later, this phenomenon can be explained by the same reason we used to explain the difference between SAFER and RanMASK in Section \ref{sec:safer}.

Note that the lower the masking rates, the more similar the class distributions RanMASK will produce for different masked copies of a given input text.
Those distributions will be extremely similar when the ``logit'' ensemble method is used, which prevents the attack algorithms from effectively identifying the weakness of victim models by applying small perturbations to the input and seeing the resulting impacts.
This misleading phenomenon deserves particular attention since it might hinder us to search for the optimal setting.
For example, as we can see from Figure \ref{fig:rate_ablation} (b), RanMASK-$5\%$ with the ``vote'' ensemble method achieved the remarkable robust accuracy on AGNEWS under three different attacks, but when the method is switched to the ``logit'' ensemble, RanMASK-$5\%$ performs much worse.
No matter which ensemble method is used, the optimal masking rate should be set to around $90\%$, which can provide the certified robustness to any perturbation on more words than the setting of $\rho = 5\%$.
As shown in Table \ref{tb-agnews-certified}, we can certify the classifications of over $50\%$ of texts to be robust to any perturbation of $5$ words on AGNEWS when $\rho = 90\%$, while the number of words allowed to be perturbed is very close to $0$ when $\rho = 5\%$.

\section{Related Work}
Even though achieving prominent performance on many important tasks, it has been reported that neural networks are vulnerable to adversarial examples---inputs generated by applying imperceptible but intentionally perturbations to examples from the dataset, such that the perturbed input can cause the model to make mistakes.
Adversarial examples were firstly discovered in the image domain \cite{szegedy2014intriguing}, and then their existence and pervasiveness were also observed in the text domain. 
Despite generating adversarial examples for texts has proven to be a more challenging task than for images due to their discrete nature, many methods have been proposed to generate adversarial text examples and reveal the vulnerability of deep neural networks in natural language processing (NLP) tasks including reading comprehension \cite{jia-liang-2017-adversarial}, text classification \cite{samanta2017towards,arxiv-17:Wong,liang2017deep, alzantot2018generating}, machine translation \cite{zhao2017generating,ebrahimi2017hotflip,cheng2020seq2sick}, dialogue systems \cite{cheng2019evaluating}, and syntactic parsing \cite{zheng2020evaluating}.
The existence and pervasiveness of adversarial examples pose serious threats to neural networks-based models especially when applying them to security-critical applications, such as face recognition systems \cite{anzhu2019generating}, autonomous driving systems \cite{eykholt2018robust}, and toxic content detection \cite{li2019textbugger}.
The introduction of the adversarial example and training ushered in a new era to understand and improve the machine learning models and has received significant attention recently \cite{goodfellow2015explaining,mohsen2016deepfool,madry2018towards, ilyas2019adversarial, pmlr-v97-cohen19c, lecuyerAG0J2019certified}. 
Adversarial examples yield broader insights into the targeted models by exposing them to such intentionally crafted examples. 
In the following, we briefly review related work in both text adversarial attacks and defenses.



\subsection{Text Adversarial Attacks}

Text adversarial attack methods generate adversarial examples by perturbing original input texts to maximally increase the model's prediction error while maintaining the adversarial examples' fluency and naturalness.
The recently proposed methods attack text
examples mainly by replacing, scrambling, and erasing characters \cite{gao2018black,ebrahimi-etal-2018-hotflip} or words \cite{alzantot2018generating,ren-etal-2019-generating, zheng2020evaluating, Jin_Jin_Zhou_Szolovits_2020, li2020bert} under the semantics- or/and syntax-preserving constraints based on the cosine similarity \cite{li2019textbugger, Jin_Jin_Zhou_Szolovits_2020}, edit distance \cite{gao2018black}, or syntactic structural similarity \cite{zheng2020evaluating, han2020adversarial}.

Depending on the degree of access to the target (or victim) model, adversarial examples can be crafted in two different settings: white-box and black-box settings \cite{arxiv-19:Xu,arxiv-19:Wang}. 
In the white-box setting, an adversary can access the model's architecture, parameters and input feature representations while not in the black-box one.
The white-box attacks normally yield a higher success rate because the knowledge of target models can be used to guide the generation of adversarial examples. 
However, the black-box attacks do not require access to target models, making them more practicable for many real-world attacks. Such attacks also can be divided into targeted and non-targeted ones depending on the purpose of adversary.
Taking classification task as an example, the output category of a generated example is intentionally controlled to a specific target category in a targeted attack while a non-targeted attack does not care about the category of misclassified results.

For text data, input sentences can be manipulated at character \cite{ebrahimi-etal-2018-hotflip}, sememe (the minimum semantic units) \cite{zang2020word}, or word \cite{samanta2017towards, alzantot2018generating} levels by replacement, alteration (e.g. deliberately introducing typos or misspellings), swap, insertion, erasure, or directly making small perturbations to their feature embeddings. Generally, we want to ensure that the crafted adversarial examples are sufficiently similar to their original ones, and these modifications should be made within semantics-preserving constraints. Such semantic similarity constraints are usually defined based on the Cosine similarity \cite{arxiv-17:Wong,barham2019interpretable,Jin_Jin_Zhou_Szolovits_2020, sears:acl18} or edit distance \cite{gao2018black}.
Zheng et al. \cite{zheng2020evaluating} show that adversarial examples also exist in syntactic parsing, and they craft the adversarial examples that preserve the same syntactic structure as the original sentences by imposing the constraints based on syntactic structural similarity.

Text adversarial example generation usually involves two steps: determine an important position (or token) to change; modify it slightly to maximize the model's prediction error. 
This two-step recipe can be repeated iteratively until the model’s prediction changes or certain stopping criteria are reached. 
Many methods have been proposed to determine the important positions 
by random selection \cite{alzantot2018generating}, trial-and-error testing at each possible point \cite{kuleshov2018adversarial}, analyzing the effects on the model of masking various parts of a input text \cite{samanta2017towards,gao2018black,Jin_Jin_Zhou_Szolovits_2020,arxiv-18:Yang}, comparing their attention scores \cite{hsieh2019robustness}, or gradient-guided optimization methods \cite{ebrahimi2017hotflip,lei2018discrete,wallace2019universal,barham2019interpretable}.

After the important positions are identified, the most popular way to alter text examples is to replace the characters or words at selected positions with similar substitutes. Such substitutes can be chosen from  nearest neighbours in an
embedding space \cite{alzantot2018generating, kuleshov2018adversarial,Jin_Jin_Zhou_Szolovits_2020,barham2019interpretable}, synonyms in a prepared dictionary \cite{samanta2017towards,hsieh2019robustness}, visually similar alternatives like typos \cite{samanta2017towards,ebrahimi2017hotflip,liang2017deep} or Internet slang and trademark logos \cite{naacl-19:Eger}, paraphrases  \cite{lei2018discrete} or even randomly selected ones \cite{gao2018black}. Given an input instance, Zhao et al. \shortcite{zhao2017generating} proposed to search for adversaries in the neighborhood of its corresponding representation in latent space by sampling within a range that is recursively tightened. Jia and Liang \shortcite{jia-liang-2017-adversarial} tried to insert few distraction sentences generated by a simple set of rules into text examples to mislead a reading comprehension system.

\subsection{Text Adversarial Defenses}
 
The goal of adversarial defenses is to learn a model capable of achieving high test accuracy on both clean and adversarial examples. 
Recently, many defense methods have been proposed to defend against text adversarial attacks, which can roughly be divided into two categories: \emph{empirical} \cite{miyato2016adversarial,sato2018interpretable,zhou2020defense,iclr-21:Dong} and \emph{certified} \cite{jia-etal-2019-certified,huang-etal-2019-achieving,ye-etal-2020-safer} methods.



Adversarial data augmentation is one of the most effective empirical defenses \cite{ren2019generating, Jin_Jin_Zhou_Szolovits_2020, li2020bert} for NLP models.
During the training time, they replace a word with one of its synonyms to create adversarial examples.
By augmenting these adversarial examples with the original training data, the model is robust to such perturbations.
However, the number of possible perturbations scales exponentially with sentence length, so data augmentation cannot cover all perturbations of an input text.
\citet{zhou2020defense} use a convex hull formed by a word and its synonyms to capture word substitution-based perturbations, and they guarantee with high probability that the model is robust at any point within the convex hull.
The similar technique also has been used by \citet{iclr-21:Dong}.
During the training phase, they allow the models to search for the worse-case over the convex hull (i.e. a set of synonyms), and minimize the error with the worst-case.
\citet{zhou2020defense} also showed that their framework can be extended to higher-order neighbors (synonyms) to boost the model's robustness further. 

Adversarial training \cite{miyato2016adversarial, zhu2019freelb} is another one of the most successful empirical defense methods by adding norm-bounded adversarial perturbations to word embeddings and minimizes the resultant adversarial loss.
A family of fast-gradient sign methods (FGSM) was introduced by  \citet{goodfellow2014explaining} to generate adversarial examples in the image domain. They showed that the robustness and generalization of machine learning models could be improved by including high-quality adversaries in the training data.
\citet{miyato2016adversarial} proposed an FGSM-like adversarial training method \cite{goodfellow2014explaining} to the text domain by applying perturbations to the word embeddings rather than to the original input itself.
\citet{sato2018interpretable} extended the work of \citet{miyato2016adversarial} to improve the interpretability by constraining the directions of perturbations toward the existing words in the word embedding space. 
\citet{arxiv-18:Zhang} applied several types of noises to perturb the input word embeddings, such as Gaussian, Bernoulli, and adversarial noises, to mitigate the overfitting problem of NLP models. 
Recently, \citet{zhu2019freelb} proposed a novel adversarial training algorithm, called FreeLB, which adds adversarial perturbations to word embeddings and minimizes the resultant adversarial loss inside different regions around input samples.
They add norm-bounded adversarial perturbations to the input sentences' embeddings using a gradient-based method and enlarge the batch size with diversified adversarial samples under such norm constraints. 
However, the studies in this line focus on the effects on generalization rather than the robustness against adversarial attacks.

Although the above empirical methods can experimentally defend the attack algorithms used during the training, 
the downside of such methods is that failure to discover an adversarial example does not mean that another more sophisticated attack could not find one. 
Recently, a set of certified defenses has been introduced, which guarantees the robustness to some specific types of attacks.
For example, Jia et al. \shortcite{jia-etal-2019-certified} and Huang et al. \shortcite{huang-etal-2019-achieving} use a bounding technique, Interval Bound Propagation \cite{arxiv-18:Gowal,arxiv-18:Dvijotham},
to formally verify a model's robustness against word substitution-based perturbations. 
\citet{iclr-20:Shi} and \citet{xu2020automatic} proposed the robustness verification and training method for transformers based on linear relaxation-based perturbation analysis.
However, these defenses often lead to loose upper bounds for arbitrary networks and result in a higher cost of clean accuracy.
Furthermore, due to the difficulty of verification, certified defense methods are usually not scale and remain hard to scale to large networks, such as BERT. 
To achieve certified robustness on large architectures, 
\citet{ye-etal-2020-safer} proposed a certified robust method, called SAFER, which is structure-free and can be applied to arbitrary large architectures.
However, the base classifier of SAFER is trained by the adversarial data augmentation, and randomly perturbing a word to its synonyms performs poorly in practice.

All existing certified defense methods make an unrealistic assumption that the defenders can access the synonyms used by the adversaries.
They could be broken by more sophisticated attacks by using synonym sets with large size \cite{Jin_Jin_Zhou_Szolovits_2020}, generating synonyms dynamically with BERT \cite{li2020bert}, or perturbing the inputs at the character-level
\cite{gao2018black,li2019textbugger}.
In this paper, we show that random smoothing can be integrating with random masking strategy to boost the certified robust accuracy.
In contrast to existing certified robust methods, the above unrealistic assumption is no longer required.
Furthermore, the NLP models trained by our defense method can defend against both the word substitution-based attacks and character-level perturbations.

This study is most related to Levine and Feizi' method \cite{levine2020robustness}, which was developed to defend against sparse adversarial attacks in the image domain.
The theoretical contribution of our study beyond the work of Levine and Feizi \cite{levine2020robustness} is that we introduce a value $\beta$ associated with each pair of an input text $\boldsymbol{x}$ and its adversarial example $\boldsymbol{x}'$, which yields a tighter certificate bound.
The value $\beta$ is defined to be the conditional probability that the base classifier $f$ will label the masked copies of $\boldsymbol{x}$ with the class $c$ where the indices of unmasked words are overlapped with $\boldsymbol{x} \ominus \boldsymbol{x}'$ (i.e., the set of word indices at which $\boldsymbol{x}$ and $\boldsymbol{x}'$ differ).
For estimating the value of $\beta$, we also present a Monte Carlo-based algorithm to evaluate $\beta$.
On MNIST \cite{deng2012mnist}, Levine and Feizi \cite{levine2020robustness} can certify the classifications of over $50\%$ of images to be robust to any perturbation of at most only $8$ pixels.
Note that the number of pixels in each image of MNIST dataset is $784$ $(28 \times 28)$, which means that just $1.02\%$ of pixels can be perturbed to provide the certified robustness.
By contrast, we can certify the classifications of over $50\%$ of texts to be robust to any perturbation of $5$ words on AGNEWS \cite{zhang2015character}.
The average length of sentences in AGNEWS dataset is about $43$, which means that $11.62\%$ of words can be maliciously manipulated while the certified robustness is guaranteed due to the tighter certificate bound provided by Equation \eqref{equ-certified-condition}.
In addition to this theoretical contribution, we proposed a new sampling strategy in which the probability of a word being masked corresponds to its output probability of a BERT-based language model (LM) to reduce the risk probability, defined a Equation \eqref{eq:probability}. 
The experimental results show that this LM-based sampling achieved better robust accuracy while suffering little loss on the clean data.

\section{Conclusion}
In this study, we propose a smoothing-based certified defense method for NLP models to substantially improve the robust accuracy against different threat models, including synonym substitution-based transformations and character-level perturbations. 
The main advantage of our method is that we do not base the certified robustness on the unrealistic assumption that the defenders know how the adversaries generate synonyms.
This method is broadly applicable, generic, scalable, and can be incorporated with little effort in any neural network, and scales to large architectures, such as BERT.
We demonstrated through extensive experimentation that our smoothed classifiers perform better than existing empirical and certified defenses across different datasets.

It would be interesting to see the results of combining RanMASK with other defense methods such as FreeLB \cite{zhu2019freelb} and DNE \cite{zhou2020defense} because they are orthogonal to each other.  
To the best of our knowledge, there does not exist a good method to boost both clean and robust accuracy, and the trade-off has been proved empirically.
We suggest a defense framework that first uses a pre-trained detector to determine whether an input text is an adversarial example, and if it is classified as an adversarial example, it will be fed to a text classifier trained with a certain defense method; otherwise, it will be input to a normally trained classifier for the prediction.
We leave these two possible improvements as future work.

\section*{Acknowledgements}
We thank Alexander Levine and Soheil Feizi \cite{levine2020robustness} for inspiring this study.
This work was supported by 
Shanghai Municipal Science and Technology Major Project (No. 2021SHZDZX0103), National Science Foundation of China (No. 62076068) and Zhangjiang Lab.

\bibliographystyle{acl_natbib}
\bibliography{acl2021}

\appendix

\section{Appendix}

\subsection{Proofs of Theorem \ref{theorem-pcx-pcxpai-diff}}
\label{sec:proof_of_theorem1}

\begingroup
\def\thetheorem{\ref{theorem-pcx-pcxpai-diff}}
\begin{theorem}
Given text $\boldsymbol{x}$ and $\boldsymbol{x}'$, $\norm{ \boldsymbol{x} - \boldsymbol{x}'}_0 \leq d$, for all class $c \in \mathcal{Y}$, we have:
\begin{equation}
 p_c(\boldsymbol{x}) - p_c(\boldsymbol{x}') \leq \beta \Delta
\end{equation}
where 
\begin{equation}
\label{equ-beta-redefinition}
\begin{gathered}
 \Delta = 1 - \frac{\tbinom{h_{\boldsymbol{x}} - d}{k_{\boldsymbol{x}}}}{\tbinom{h_{\boldsymbol{x}}}{k_{\boldsymbol{x}}}}, 
 \\
 \beta = \mathbb{P}(f(\mathcal{M}(\boldsymbol{x}, \mathcal{H})) = c \mid \mathcal{H} \cap (\boldsymbol{x} \ominus \boldsymbol{x}') \neq \emptyset).
 \end{gathered}
\end{equation}
\end{theorem}
\addtocounter{theorem}{-1}
\endgroup

\begin{proof}
Recall that $\mathcal{H} \sim \mathcal{U}(h_{\boldsymbol{x}}, k_{\boldsymbol{x}})$, we have:
\begin{equation}
\begin{gathered}
 p_c(\boldsymbol{x}) = \mathbb{P} (f(\mathcal{M}(\boldsymbol{x}, \mathcal{H})) = c),
 \\
 p_c(\boldsymbol{x}') = \mathbb{P} (f(\mathcal{M}(\boldsymbol{x}', \mathcal{H})) = c).
 \end{gathered}
\end{equation}

\noindent By the law of total probability, we have:
\begin{equation}
\label{equ-pcx-pcxpai}
\begin{aligned}
 p_c(\boldsymbol{x}) = & \mathbb{P} ([f(\mathcal{M}(\boldsymbol{x}, \mathcal{H})) = c] \wedge [\mathcal{H} \cap (\boldsymbol{x} \ominus \boldsymbol{x}') = \emptyset]) +
 \\ 
 & \mathbb{P} ([f(\mathcal{M}(\boldsymbol{x}, \mathcal{H})) = c] \wedge [\mathcal{H} \cap (\boldsymbol{x} \ominus \boldsymbol{x}') \neq \emptyset]) 
 \\
 p_c(\boldsymbol{x}') = & \mathbb{P} ([f(\mathcal{M}(\boldsymbol{x}', \mathcal{H})) = c] \wedge [\mathcal{H} \cap (\boldsymbol{x} \ominus \boldsymbol{x}') = \emptyset ]) +
 \\
 & \mathbb{P} ([f(\mathcal{M}(\boldsymbol{x}', \mathcal{H})) = c] \wedge [\mathcal{H} \cap (\boldsymbol{x} \ominus \boldsymbol{x}') \neq \emptyset]).
 \end{aligned}
\end{equation}

\noindent 
Note that if $\mathcal{H} \cap (\boldsymbol{x} \ominus \boldsymbol{x}') = \emptyset$, then $\boldsymbol{x}$ and $\boldsymbol{x}'$ are identical at all indices in $\mathcal{H}$.
In this case, we have $\mathcal{M}(\boldsymbol{x},\mathcal{H}) = \mathcal{M}(\boldsymbol{x}',\mathcal{H})$, which implies:
\begin{equation}
\label{equ-equal-conditional-sample-empty}
\begin{aligned}
\mathbb{P}(f(\mathcal{M}(\boldsymbol{x}, \mathcal{H})) = c \mid \mathcal{H} \cap (\boldsymbol{x} \ominus \boldsymbol{x}') = \emptyset) & = 
\\
\mathbb{P}(f(\mathcal{M}(\boldsymbol{x}', \mathcal{H})) = c \mid \mathcal{H} \cap (\boldsymbol{x} \ominus \boldsymbol{x}') = \emptyset) &
 \end{aligned}
\end{equation}

\noindent Multiplying both sizes of (\ref{equ-equal-conditional-sample-empty}) by $\mathbb{P} (\mathcal{H} \cap (\boldsymbol{x} \ominus \boldsymbol{x}') = \emptyset)$, we obtain: 
\begin{equation}
\label{equ-equal-allprob-sample-empty}
\begin{aligned}
\mathbb{P} ([f(\mathcal{M}(\boldsymbol{x}, \mathcal{H})) = c] \wedge [\mathcal{H} \cap (\boldsymbol{x} \ominus \boldsymbol{x}') = \emptyset]) & = 
\\
\mathbb{P} ([f(\mathcal{M}(\boldsymbol{x}', \mathcal{H})) = c] \wedge [\mathcal{H} \cap (\boldsymbol{x} \ominus \boldsymbol{x}') = \emptyset]) &
 \end{aligned}
\end{equation}

\noindent By (\ref{equ-pcx-pcxpai}), (\ref{equ-equal-allprob-sample-empty}) and the non-negativity of probability, subtracting $p_c(\boldsymbol{x}')$ from $p_c(\boldsymbol{x})$ yields:
\begin{equation}
\label{equ-subtraction}
\begin{aligned}
 p_c(&\boldsymbol{x}) - p_c(\boldsymbol{x}') = 
\\ 
& \mathbb{P} ([f(\mathcal{M}(\boldsymbol{x}, \mathcal{H})) = c] \wedge [\mathcal{H} \cap (\boldsymbol{x} \ominus \boldsymbol{x}') \neq \emptyset]) - 
\\
& \mathbb{P} ([f(\mathcal{M}(\boldsymbol{x}', \mathcal{H})) = c] \wedge [\mathcal{H} \cap (\boldsymbol{x} \ominus \boldsymbol{x}') \neq \emptyset])
\\
\leq & \mathbb{P} ([f(\mathcal{M}(\boldsymbol{x}, \mathcal{H})) = c] \wedge [\mathcal{H} \cap (\boldsymbol{x} \ominus \boldsymbol{x}') \neq \emptyset])
 \end{aligned}
\end{equation}

\noindent By the definition of $\beta$, we have 
\begin{equation}
\label{equ-beta}
\begin{aligned}
\mathbb{P} ([f(\mathcal{M}(\boldsymbol{x}, \mathcal{H})) = c] \wedge [\mathcal{H} \cap (\boldsymbol{x} \ominus \boldsymbol{x}') \neq \emptyset]) & = \\ \beta \times \mathbb{P} (\mathcal{H} \cap (\boldsymbol{x} \ominus \boldsymbol{x}') \neq \emptyset) &
\end{aligned}
\end{equation}
\noindent Substituting (\ref{equ-beta}) into (\ref{equ-subtraction}) gives:
\begin{equation}
\label{equ-pcx-pcxpai-inequality}
p_c(\boldsymbol{x}) - p_c(\boldsymbol{x}') \leq \beta \times \mathbb{P} (\mathcal{H} \cap (\boldsymbol{x} \ominus \boldsymbol{x}') \neq \emptyset)
\end{equation}
Note that,
\begin{equation}
\mathbb{P} (\mathcal{H} \cap (\boldsymbol{x} \ominus \boldsymbol{x}') = \emptyset) = \frac{\tbinom{h_{\boldsymbol{x}} - |\boldsymbol{x} \ominus \boldsymbol{x}'|}{k_{\boldsymbol{x}}}}{\tbinom{h_{\boldsymbol{x}}}{k_{\boldsymbol{x}}}} = \frac{\tbinom{h_{\boldsymbol{x}} - \norm{\boldsymbol{x} - \boldsymbol{x}'}_0}{k_{\boldsymbol{x}}}}{\tbinom{h_{\boldsymbol{x}}}{k_{\boldsymbol{x}}}}
\end{equation}

\noindent where the last equality follows since $\mathcal{H}$ is a uniform choice of $k_{\boldsymbol{x}}$ elements from $h_{\boldsymbol{x}}$: there are $\tbinom{h_{\boldsymbol{x}}}{k_{\boldsymbol{x}}}$ total ways to make this selection. Among all these selections, there are $\tbinom{h_{\boldsymbol{x}} - |\boldsymbol{x} \ominus \boldsymbol{x}'|}{k_{\boldsymbol{x}}}$ of which do not contain any element from $\boldsymbol{x} \ominus \boldsymbol{x}'$.

\noindent By the constraint of $\norm{ \boldsymbol{x} - \boldsymbol{x}'}_0 \leq d$, we have:
\begin{equation}
\label{equ-delta-inequality}
\begin{aligned}
\mathbb{P} (\mathcal{H} \cap (\boldsymbol{x} \ominus \boldsymbol{x}') \neq \emptyset) & = 
1 - \mathbb{P} (\mathcal{H} \cap (\boldsymbol{x} \ominus \boldsymbol{x}') = \emptyset) 
\\
& = 1 -\frac{\tbinom{h_{\boldsymbol{x}} - \norm{\boldsymbol{x} - \boldsymbol{x}'}_0}{k_{\boldsymbol{x}}}}{\tbinom{h_{\boldsymbol{x}}}{k_{\boldsymbol{x}}}}
\\
& \leq 1 - \frac{\tbinom{h_{\boldsymbol{x}} - d}{k_{\boldsymbol{x}}}}{\tbinom{h_{\boldsymbol{x}}}{k_{\boldsymbol{x}}}} = \Delta
 \end{aligned}
\end{equation}

\noindent Combining inequalities (\ref{equ-pcx-pcxpai-inequality}) and (\ref{equ-delta-inequality}) gives the statement of Theorem \ref{theorem-pcx-pcxpai-diff}.
\end{proof}

\end{CJK}
\end{document}